\def\paperlanguage{}
\pdfoutput=1 

\documentclass[letterpaper, 10 pt, conference]{ieeeconf}  

\usepackage{bm}
\usepackage{cite}
\usepackage{flushend}
\include{preamble}

\IEEEoverridecommandlockouts                              

\title{\LARGE \textbf
  {
    \switchlanguage%
    {%
      Deep Predictive Model Learning with Parametric Bias:\\Handling Modeling Difficulties and Temporal Model Changes
    }%
    {%
      Parametric Biasを含む深層予測モデル学習:\\モデル化困難性と逐次的モデル変化への対処
    }%
  }
}

\author{Kento Kawaharazuka$^{1}$, Kei Okada$^{1}$, and Masayuki Inaba$^{1}$
  \thanks{$^{1}$ The authors are with the Department of Mechano-Informatics, Graduate School of Information Science and Technology, The University of Tokyo, 7-3-1 Hongo, Bunkyo-ku, Tokyo, 113-8656, Japan.
    {\texttt\small [kawaharazuka, k-okada, inaba]@jsk.t.u-tokyo.ac.jp}
  }
}
\begin{document}

\maketitle
\thispagestyle{empty}
\pagestyle{empty}

\begin{abstract}
  \switchlanguage%
  {%
    When a robot executes a task, it is necessary to model the relationship among its body, target objects, tools, and environment, and to control its body to realize the target state.
    However, it is difficult to model them using classical methods if the relationship is complex.
    In addition, when the relationship changes with time, it is necessary to deal with the temporal changes of the model.
    In this study, we have developed Deep Predictive Model with Parametric Bias (DPMPB) as a more human-like adaptive intelligence to deal with these modeling difficulties and temporal model changes.
    We categorize and summarize the theory of DPMPB and various task experiments on the actual robots, and discuss the effectiveness of DPMPB.
  }%
  {%
    ロボットがタスクを実行する際, その身体, 対象物体, 道具, 動作環境の間の関係性をモデル化し, 指令値を実現するように身体を制御する必要がある.
    しかし, それらの関係が複雑である場合, 古典的手法ではそのモデル化自体が難しい.
    また, それらの関係が時間と伴に変化する場合, そのモデルの逐次的変化に対応する必要がある.
    そこで本研究では, このモデル化困難性と逐次的モデル変化に対処するより人間らしい適応的知能として, Deep Predictive Model with Parametric Bias (DPMPB)を開発してきた.
    その理論と様々なロボット・タスクにおける実験を分類しまとめ, その有効性について考察する.
  }%
\end{abstract}

\section{INTRODUCTION}\label{sec:introduction}
\switchlanguage%
{%
  The purpose of a robot is to realize the target task state when given a certain task by manipulating its own body, target objects, and tools.
  This is possible by modeling and applying the relationships among its body, target objects, tools, and environment.
  For example, cloth manipulation can be achieved by knowing how to move the arm to change the state of the grasped cloth.
  Also, balance control can be achieved by knowing how to apply force to the ankle to change the zero moment point.
  However, there are two major problems here: modeling difficulties and temporal model changes.

  First, when the relationship among the body, tools, target objects, and environment is complex, it is difficult to model using classical methods.
  Most classical methods in the past have dealt with robots that are rigid and easy to model, or tools and objects that can be approximated by rigid bodies \cite{kobayashi1998tendon, kemp2006tooltip}.
  However, nowadays, robots with flexible bodies \cite{lee2017softrobotics} and tasks that handle flexible objects \cite{tanaka2018emd} are attracting attention, and it is necessary to develop methods that can be applied to them.
  Second, when the relationship among the body, tools, target objects, and environment changes with time, it is necessary to deal with the temporal changes in the model.
  In the past, target objects, tools, and environment were mostly fixed, but now robots are expected to be deployed in more complex environments where they change dynamically \cite{kuniyoshi2004chaos, lee2020anymal}.
  In addition, the nature of flexible robots makes them prone to physical changes due to aging and other factors, and they should be able to cope with the temporal model changes caused by these physical changes.
  Robots need a learning system resembling human adaptive intelligence that allows them to cope with these problems in the real world.
}%
{%
  ロボットは一般的に, タスクを与えられた際にその指令状態を実現するように制御される.
  このとき, その身体, 対象物体, 道具, 動作環境の間の関係性をモデル化し, これを応用することでタスクが実行される.
  例えば, どう腕を動かすと把持した布の状態が変化するかや, どう足首に力を入れると圧力中心が変化するかを知ることによって, マニピュレーションやバランス制御が可能となる.
  しかし, ここにはモデル化困難性と逐次的モデル変化という大きな2つの問題がある.

  まず, 身体-道具-対象物体-動作環境の関係が複雑である場合, 古典的手法ではそのモデル化自体が難しい.
  これまでは高剛性でモデル化の容易なロボットや剛体近似可能な道具, 対象物体を扱う場合がほとんどであった\cite{kobayashi1998tendon, kemp2006tooltip}.
  しかし, 現在は柔軟な身体を持つロボット\cite{lee2017softrobotics}や柔軟物体を扱うタスク\cite{tanaka2018emd}に対しても目が向けられつつあり, これらに適用可能な手法を開発する必要がある.
  次に, 身体-道具-対象物体-動作環境の関係が時間と伴に変化する場合, そのモデルの逐次的変化に対応する必要がある.
  これまでは対象物体や道具, 動作環境が変化しない場合がほとんどであったが, 現在はこれらが動的に変化するより複雑な環境\cite{kuniyoshi2004chaos, lee2020anymal}へのロボットの投入が期待されている.
  また, 柔軟なロボットはその性質から経年劣化等による身体変化が起きやすく, これらによる逐次的モデル変化にも対応するべきである.
  実世界でこれらの問題点に柔軟に対応可能な, 人間の適応的知能のような学習システムが必要である.
}%

\begin{figure}[t]
  \centering
  \includegraphics[width=0.99\columnwidth]{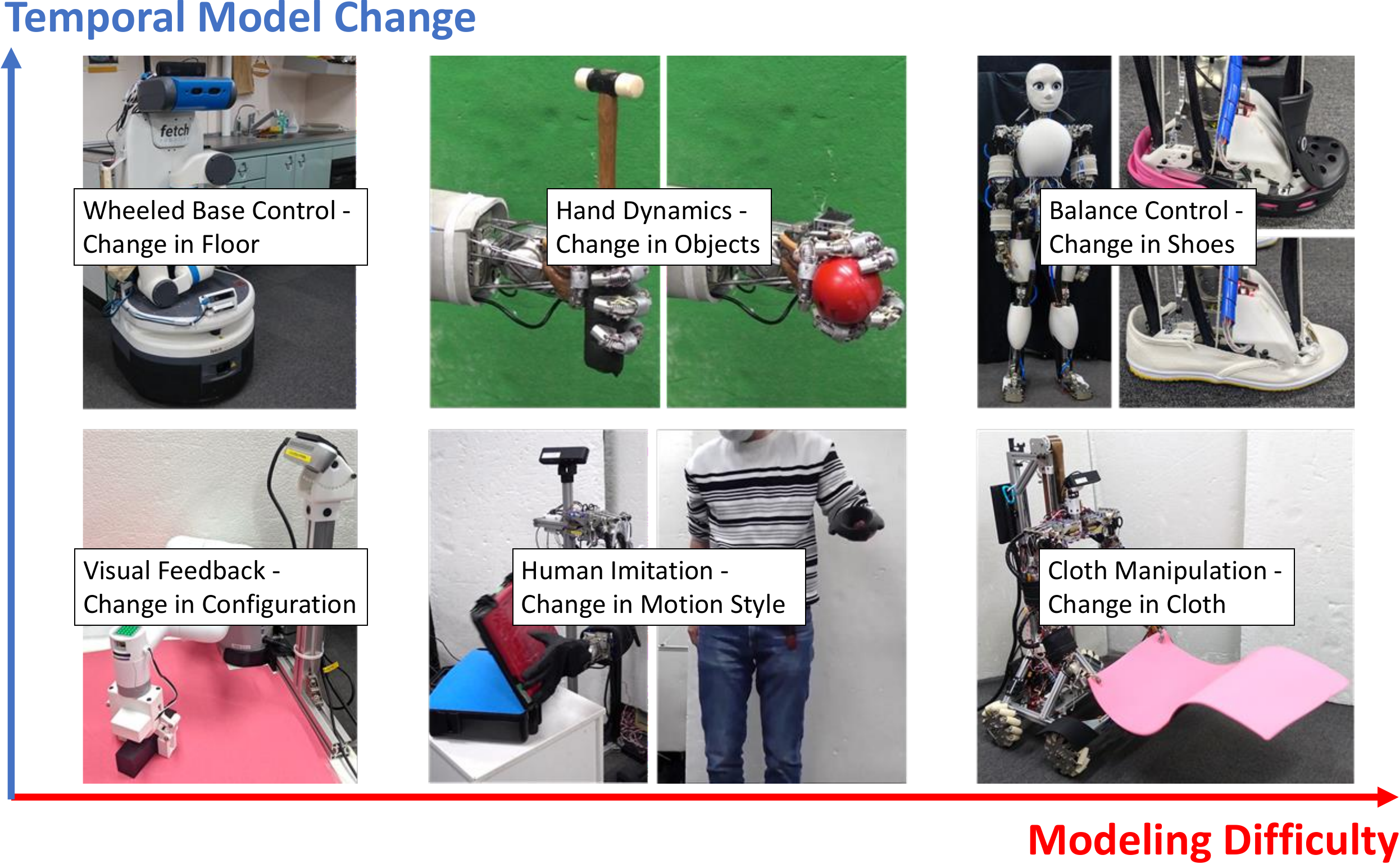}
  \vspace{-3.0ex}
  \caption{The developed deep predictive model with parametric bias (DPMPB) can handle various modeling difficulties and temporal model changes.}
  \label{figure:concept}
  \vspace{-3.0ex}
\end{figure}

\begin{figure*}[t]
  \centering
  \includegraphics[width=2.0\columnwidth]{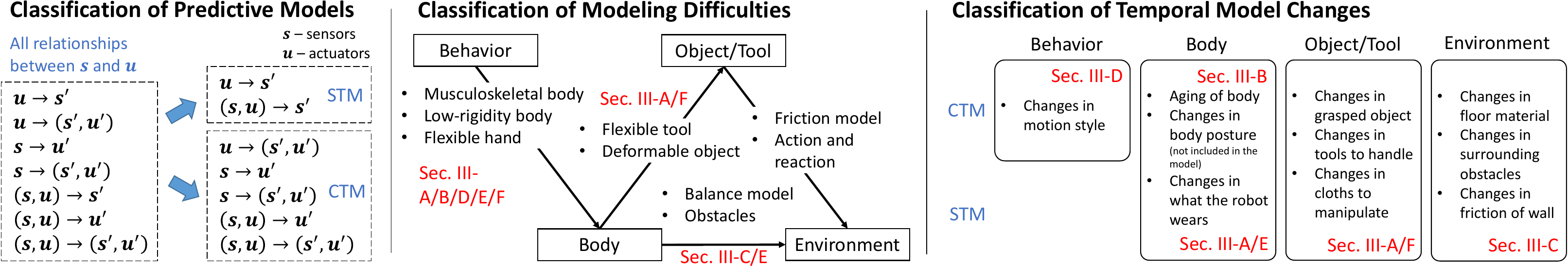}
  \vspace{-1.0ex}
  \caption{The classification of predictive models, modeling difficulties, and temporal model changes. The predictive models are classified by the network input and output of the values of sensors $\bm{s}$ and actuators $\bm{u}$. The modeling difficulties are classified by the relationship among robot behavior, body, object/tool, and environment. The temporal model changes are classified by the network structure of CTM or STM and by robot behavior, body, object/tool, and environment.}
  \label{figure:classification}
  \vspace{-3.0ex}
\end{figure*}

\switchlanguage%
{%
  Therefore, we have developed Deep Predictive Model with Parametric Bias (DPMPB) to cope with these modeling difficulties and temporal model changes \cite{kawaharazuka2020dynamics, kawaharazuka2022vservoing, kawaharazuka2021fetch, kawaharazuka2021imitation, kawaharazuka2022balance, kawaharazuka2022cloth}.
  This predictive model describes the correlation between sensors $\bm{s}$ and actuators $\bm{u}$ with a neural network.
  This learning-based modeling makes it possible to cope with modeling difficulties.
  By using this predictive model, the robot body is controlled by its forward propagation or by iterative backpropagation from an appropriate loss function to the network input.
  It is also possible to detect anomalies based on the prediction error of the model.

  In order to cope with temporal model changes in such learning-based modeling, we apply parametric bias \cite{tani2002parametric}, which can implicitly embed multiple attractor dynamics.
  Parametric bias is a learnable input variable, and dynamics information is embedded in this variable so that various data transitions with different dynamics can be represented in a single model.
  This has been used mainly in the context of cognitive robotics with imitation learning \cite{ogata2005extracting}.
  By applying this approach to predictive model learning, it is possible to implicitly embed temporal changes in the body, tools, target objects, and environment.
  In addition, unlike \cite{tani2002parametric}, we develop a mechanism that autonomously updates the parametric bias online, after which the control changes accordingly.
  This will enable the system to recognize and adapt to the current state of the body, tools, target objects, and environment.

  Reinforcement learning \cite{tedrake2004stochastic} and imitation learning \cite{zhang2018imitation} are commonly considered as control methods to deal with modeling difficulties.
  Reinforcement learning is a method that can acquire controllers through autonomous learning based on rewards.
  While it is mainly applied to simulations where the number of trials can be increased, efficient reinforcement learning methods that can be learned on actual robots have also been developed \cite{yang2019legged}.
  Imitation learning is a method where a robot learns to imitate a human demonstration, and is a type of predictive model that is handled in this study.
  In addition, a predictive model representing state transition has been developed \cite{schemeckpeper2020predictive}, which is also handled in this study.
  On the other hand, there are very few cases in which the problem of temporal model changes is explicitly addressed in the learning methods that deal with these modeling difficulties.
  All of reinforcement learning studies deal with the temporal model changes implicitly by learning in various environments, but few of them deal with the temporal changes of bodies, objects, and tools.
  Predictive model learning studies so far basically do not deal with the changes in bodies, tools, and environments \cite{kawaharazuka2019dynamic}, and the robot is controlled with the fixed dynamics from the time it was trained.
  There is also a method using online learning of neural networks to adapt to the current body and environment \cite{kawaharazuka2020autoencoder}, but it requires a large amount of data to relearn the entire network, losing its applicability to other bodies and environments.
  By introducing parametric bias, changes in the body and environment can be embedded in small dimensional variables, which can be updated online to adapt to the current body, tool, and environment quickly without destroying the dynamics of the entire network.
  In addition, since the changes in the body and environment can be taken into account in the model as explicit variables, they can be applied to the recognition of the body and environment, thus expanding the range of possible tasks.

  In this study, we introduce parametric bias in predictive model learning, and discuss how to cope with the modeling difficulties and temporal model changes based on this approach.
  We summarize the actual robot examples using predictive model learning that we have developed so far \cite{kawaharazuka2020dynamics, kawaharazuka2022vservoing, kawaharazuka2021fetch, kawaharazuka2021imitation, kawaharazuka2022balance, kawaharazuka2022cloth}, and integrate the methods into a single unified theory, DPMPB.
  The predictive model learning is classified into state transition model type and control transition model type, and these types are identified with the collected dataset.
  Modeling difficulties and temporal model changes in the predictive model learning are classified and summarized.
  The theory of DPMPB includes data collection, network training, online adaptation, control, and anomaly detection in the form of forward and backward propagation of inputs and outputs of the constructed network.
  We also describe a concrete implementation of our system with necessary parameters specified.
  Based on this unified theory, we show that experiments on actual robot tasks with various modeling difficulties and temporal model changes can be comprehensively achieved by simply changing parameters and training data in exactly the same form (\figref{figure:concept}).
}%
{%
  そこで本研究では, このモデル化困難性と逐次的モデル変化に対応する, Deep Predictive Model with Parametric Bias (DPMPB)を開発してきた\cite{kawaharazuka2020dynamics, kawaharazuka2022vservoing, kawaharazuka2021fetch, kawaharazuka2021imitation, kawaharazuka2022balance, kawaharazuka2022cloth}.
  この予測モデルは感覚$\bm{s}$と運動$\bm{u}$の相関関係をニューラルネットワークにより記述する.
  この学習型のモデル化により, モデル化困難性に対応することが可能になる.
  この予測モデルを用いて, その順伝播, または設定した損失関数からネットワーク入力に対する繰り返しの誤差逆伝播により身体を制御する.
  予測モデルの誤差に基づき異常検知も可能である.

  そして, このような学習型のモデル化において, 逐次的モデル変化に対応するために, 複数のアトラクターダイナミクスを暗黙的に埋め込むことが可能なParametric Bias \cite{tani2002parametric}を応用する.
  Parametric Biasは学習可能な入力変数であり, 異なるダイナミクスを持つ様々なデータ遷移を一つのモデルで表現できるようネットワークの入力変数の一部にこのダイナミクス情報を埋め込む.
  これは, これまで主に認知ロボティクスや模倣学習の文脈で利用されてきた\cite{ogata2005extracting}.
  これを予測モデル学習に応用することで, 身体-道具-対象物体-動作環境の逐次的変化を暗黙的に埋め込むことが可能になる.
  また, \cite{tani2002parametric}とは異なり, Parametric Biasをオンラインで自律的に更新していき, それに応じて制御が変化していく機構を開発する.
  これによって, 現在の身体-道具-対象物体-動作環境状態を認識し, 適応することができるようになる.

  モデル化困難性に対応するロボットの動作手法として, 強化学習\cite{tedrake2004stochastic}と模倣学習\cite{zhang2018imitation}が一般的に考えられる.
  強化学習は報酬に基づき自律的な学習により制御器を獲得することができる手法である.
  基本的に試行回数を稼げるシミュレーションにおける適用が主であるが, 実機のみで実行可能な効率的な強化学習手法も開発されてきている\cite{yang2019legged}.
  模倣学習は人間によるデモンストレーションからそれを模倣するようにロボットが感覚と運動の遷移を学習する手法であり, 本研究でも扱う予測モデルの一種である.
  加えて, \cite{schemeckpeper2020predictive}のような状態遷移型の予測モデルも本研究で扱う.
  一方で, これらのモデル化困難性に対応する様々な学習手法において, 逐次的モデル変化という問題点を明示的に扱っているケースは非常に少ない.
  強化学習はそのどれもが, 様々な環境において学習を行うことでそれらに対応可能な方策を獲得する形であり, 特に身体や道具の変化等を考慮した研究はほとんどない.
  予測モデル学習は基本的に身体や道具, 環境の変化を扱わず, ダイナミクスは学習された時から固定として制御を行っている.
  また, 現在の身体や環境に適応するためにニューラルネットワークのオンライン学習を適用する例もあるが, ネットワーク全体を学習し直すには多くのデータが必要かつ他の身体や環境への適用性を失ってしまう.
  Parametric Biasを導入することで, 身体や環境の変化を小さな次元の変数に埋め込み, これのみをオンラインで更新することで, 素早くかつネットワーク全体のダイナミクスを崩さずに現在の身体や道具, 環境に適応することができる.
  加えて, 明示的な変数としてその身体や環境の変化をモデル内に考慮できるようになるため, これを逆に身体や環境の認識に応用することができ, 可能となるタスクの幅も広がる.

  本研究では, 予測モデル学習におけるParametric Biasの導入と, これに基づくロボットのモデル化困難性と逐次的モデル変化への対処について述べる.
  これまで開発した予測モデル学習の実例\cite{kawaharazuka2020dynamics, kawaharazuka2022vservoing, kawaharazuka2021fetch, kawaharazuka2021imitation, kawaharazuka2022balance, kawaharazuka2022cloth}について, その手法をDPMPBという形で理論として統合しまとめる.
  予測モデル学習を状態遷移モデル型と運動遷移モデル型に分けこれを同定する.
  また, 予測モデル学習におけるモデル化困難性と逐次的モデル変化を分類しまとめる.
  DPMPBの理論は, 構築したネットワークの入出力の順伝播と逆伝播を駆使した形で, データ収集, ネットワーク学習, オンライン適応, 制御, 異常検知を含んでいる.
  必要なパラメータ等を明示した具体的な実装方法についても述べる.
  この統一理論をもとに, 全く同じ形式でパラメータやデータを変更するだけで, 様々なモデル化困難性と逐次的モデル変化を伴うロボット・タスクにおける実験が網羅的に達成できることを示す(\figref{figure:concept}).
}%

\begin{figure*}[t]
  \centering
  \includegraphics[width=1.7\columnwidth]{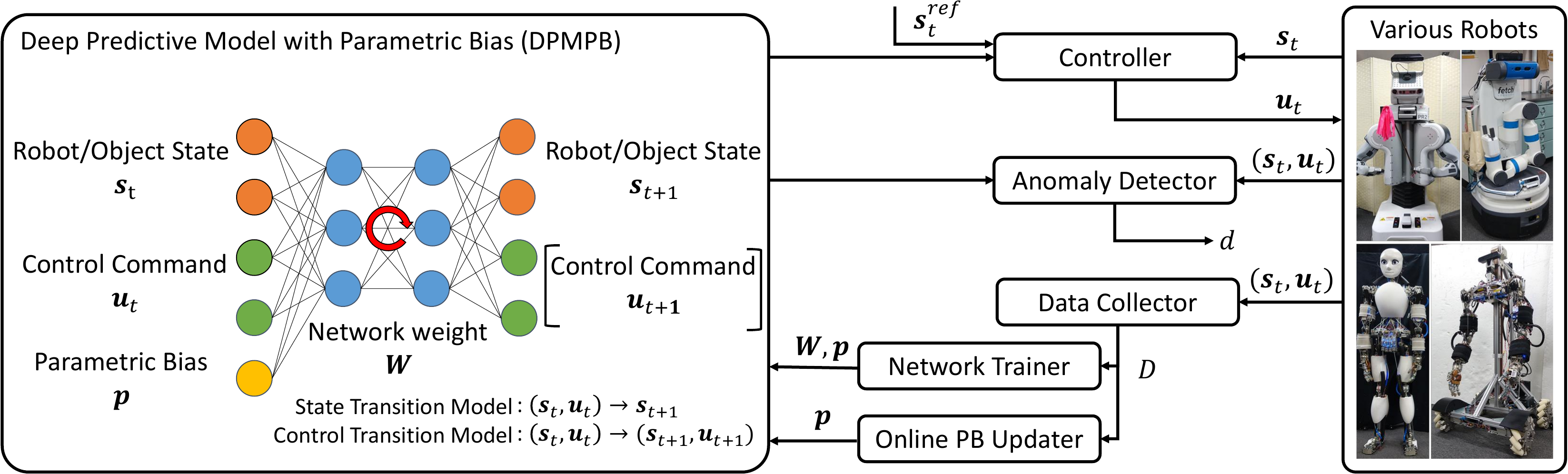}
  \vspace{-1.0ex}
  \caption{The system overview of deep predictive model with parametric bias (DPMPB). DPMPB has the network input of robot/object state $\bm{s}_{t}$, control command $\bm{u}_{t}$, and parametric bias $\bm{p}$, and the network output of $\bm{s}_{t+1}$ and $\bm{u}_{t+1}$ depending on the network structure of state transition model (STM) or control transition model (CTM). Controller, Anomaly Detector, Data Collector, Network Trainer, and Online PB Updater of various robots can be executed through DPMPB by only changing the network input/output and a few parameters.}
  \label{figure:network-structure}
  \vspace{-3.0ex}
\end{figure*}

\section{Deep Predictive Model with Parametric Bias} \label{sec:proposed}

\subsection{Classification of Predictive Models, Modeling Difficulties, and Temporal Model Changes} \label{subsec:classification}
\switchlanguage%
{%
  \subsubsection{Classification of Deep Predictive Models}

  First, we classify the predictive models based on the relationship between sensors $\bm{s}$ and actuators $\bm{u}$ (the left figure of \figref{figure:classification}).
  There are seven possible time evolution relationships between $\bm{s}$ and $\bm{u}$: $\{\bm{u}\rightarrow\bm{s}', \bm{u}\rightarrow(\bm{s}', \bm{u}'), \bm{s}\rightarrow\bm{u}', \bm{s}\rightarrow(\bm{s}', \bm{u}'), (\bm{s}, \bm{u})\rightarrow\bm{s}', (\bm{s}, \bm{u})\rightarrow\bm{u}', (\bm{s}, \bm{u})\rightarrow(\bm{s}', \bm{u}')\}$ ($\{\bm{s}', \bm{u}'\}$ represents $\{\bm{s}, \bm{u}\}$ at the next time step).
  The five that contain $\bm{u}$ in the output, $\{\bm{u}\rightarrow(\bm{s}', \bm{u}'), \bm{s}\rightarrow\bm{u}', \bm{s}\rightarrow(\bm{s}', \bm{u}'), (\bm{s}, \bm{u})\rightarrow\bm{u}', (\bm{s}, \bm{u})\rightarrow(\bm{s}', \bm{u}')\}$, have the network structure of control transition model (CTM) used in imitation learning, in which the motion of the next time step is output from the current state.
  Also, since $\bm{u}\rightarrow\bm{s}'$ and $(\bm{s}, \bm{u})\rightarrow\bm{s}'$ contain only $\bm{s}$ in the output, they can be regarded as the network structure of state transition model (STM) that predicts the state of the next time step.
  In CTM, $(\bm{s}, \bm{u})\rightarrow(\bm{s}', \bm{u}')$ contains all the information of the other four network structures, and in STM, $(\bm{s}, \bm{u})\rightarrow\bm{s}'$ contains all the information of the other network structure.
  The predictive model learning with the network structures of $(\bm{s}, \bm{u})\rightarrow\bm{s}'$ (STM) and $(\bm{s}, \bm{u})\rightarrow(\bm{s}', \bm{u}')$ (CTM) are handled in this study, since these two network structures can show the basic functions of DPMPB.

  \subsubsection{Classification of Modeling Difficulties}

  Next, we classify the modeling difficulties handled in this study (the center figure of \figref{figure:classification}).
  The behavior of a robot propagates its effects in the form of its body, the object/tool it handles, and the environment in which it operates.
  For this reason, four categories are defined: behavior, body, object/tool, and environment.
  Predictive models represent the relationships among multiple sensors and actuators.
  Therefore, modeling difficulties can be categorized into the modeling between behavior and body, between body and object/tool, between body and environment, and between object/tool and environment.
  The modeling difficulty between behavior and body occur when the body is complex, especially with flexible and redundant musculoskeletal systems, low-rigidity plastic bodies, flexible hands, etc. (handled in \secref{subsec:hand-dynamics}, \secref{subsec:visual-feedback}, \secref{subsec:imitation-learning}, \secref{subsec:balance-control}, and \secref{subsec:cloth-manipulation}).
  The modeling difficulty between body and object/tool occurs when the robot body handles flexible tools or flexible objects (handled in \secref{subsec:hand-dynamics} and \secref{subsec:cloth-manipulation}).
  The modeling difficulty between body and environment occurs when the body interacts with various undefined environments such as unknown floor material and obstacles (handled in \secref{subsec:stable-fetch} and \secref{subsec:balance-control}).
  Although not directly handled in this study, modeling difficulties between object/tool and environment, such as friction and action-reaction, are also likely to occur.

  \subsubsection{Classification of Temporal Model Changes}

  Finally, we classify the temporal model changes handled in this study (the right figure of \figref{figure:classification}).
  We can classify them by STM or CTM, and also by the robot behavior, body, object/tool, or environment.
  Here, temporal model changes can be detected because the model can predict the changing values of sensors and actuators.
  Therefore, CTM can detect temporal model changes for behavior, body, object/tool, and environment, but STM cannot consider temporal model changes for behavior because it does not predict behavior.
  We will raise examples of temporal model changes.
  Regarding behavior, we can consider the changes in behavior for the same state, i.e., the change in motion style (handled in \secref{subsec:imitation-learning}).
  Regarding body, we can consider the changes in body state due to aging, changes in parts of the body not directly included in the model, changes in what the robot wears, etc. (handled in \secref{subsec:visual-feedback} for CTM, and in \secref{subsec:hand-dynamics} and \secref{subsec:balance-control} for STM).
  For object/tool, we can consider changes in grasped objects, handled tools, manipulated cloths, etc. (handled in \secref{subsec:hand-dynamics} and \secref{subsec:cloth-manipulation}).
  For environment, we can consider changes in floor materials, surrounding obstacles, wall friction, etc. (handled in \secref{subsec:stable-fetch}).

  DPMPB can realize various tasks by combining its structure (STM or CTM), the handled modeling difficulties among behavior, body, object/tool, and environment, and the handled temporal model changes in behavior, body, object/tool, and environment.
  Specifically, this combination corresponds to changes in the parameters of whether $\bm{u}$ is included in the network output, what $\bm{s}$ and $\bm{u}$ are used, and what changes occur when $\bm{s}$ and $\bm{u}$ are collected.
  In this study, we show that it is possible to realize a comprehensive set of possible task examples based on these classifications by simply changing these parameters in DPMPB.
}%
{%
  \subsubsection{Classification of Deep Predictive Models}

  まず, 感覚$\bm{s}$と運動$\bm{u}$の関係から予測モデルを分類する(\figref{figure:classification}の左図).
  $\bm{s}$と$\bm{u}$の時間発展関係には, $\{\bm{u}\rightarrow\bm{s}', \bm{u}\rightarrow(\bm{s}', \bm{u}'), \bm{s}\rightarrow\bm{u}', \bm{s}\rightarrow(\bm{s}', \bm{u}'), (\bm{s}, \bm{u})\rightarrow\bm{s}', (\bm{s}, \bm{u})\rightarrow\bm{u}', (\bm{s}, \bm{u})\rightarrow(\bm{s}', \bm{u}')\}$の7つが考えられる($\{\bm{s}, \bm{u}\}'$は次時刻の$\{\bm{s}, \bm{u}\}$を表す).
  このうち, $\{\bm{u}\rightarrow(\bm{s}', \bm{u}'), \bm{s}\rightarrow\bm{u}', \bm{s}\rightarrow(\bm{s}', \bm{u}'), (\bm{s}, \bm{u})\rightarrow\bm{u}', (\bm{s}, \bm{u})\rightarrow(\bm{s}', \bm{u}')\}$の5つは出力に$\bm{u}$を含むため, 現在感覚から次時刻の運動を出力するような, 模倣学習で用いられる運動遷移モデル型のネットワーク構造(Control Transition Model, CTM)となる.
  また, $\bm{u}\rightarrow\bm{s}'$と$(\bm{s}, \bm{u})\rightarrow\bm{s}'$は出力に$\bm{s}$のみを含むため, 感覚状態を予測する状態遷移モデル型のネットワーク構造(State Transition Model, STM)と言える.
  CTMにおいて, $(\bm{s}, \bm{u})\rightarrow(\bm{s}', \bm{u}')$は他4つのネットワーク構造の情報を全て含み, STMにおいて, $(\bm{s}, \bm{u})\rightarrow\bm{s}'$は他1つのネットワーク構造の情報を全て含む.
  この2種類により予測モデルの基本要素は示せるため, $(\bm{s}_{t}, \bm{u}_{t})\rightarrow\bm{s}_{t+1}$ (STM)と$(\bm{s}_{t}, \bm{u}_{t})\rightarrow(\bm{s}_{t+1}, \bm{u}_{t+1})$ (CTM)のネットワーク構造を持つ予測モデル学習を本研究では扱う.

  \subsubsection{Classification of Modeling Difficulties}

  次に, 本研究で扱うモデル化困難性を分類する(\figref{figure:classification}の中図).
  ロボットの行動は, 行動によって動作する身体, その身体が扱う物体や道具, そして動作する環境という形でその影響が伝播する.
  そのため, 行動, 身体, 物体/道具, 環境という4つのカテゴリを設けた.
  予測モデルは, 複数の感覚や運動の相関関係を表現するものである.
  ゆえに, モデル化困難性は, 行動と身体の間のモデリング, 身体と物体/道具の間のモデリング, 身体と環境の間のモデリング, 物体/道具と環境の間のモデリングに分類することができる.
  行動と身体の間のモデル化困難性は, 身体が複雑である場合, とりわけ, 柔軟で冗長な筋骨格系や低剛性樹脂製の身体, 柔軟ハンド等に起こる(これらは\secref{subsec:hand-dynamics}, \secref{subsec:visual-feedback}, \secref{subsec:imitation-learning}, \secref{subsec:balance-control}, \secref{subsec:cloth-manipulation}で扱う).
  また, 身体と物体/道具の間のモデル化困難性は, 柔軟な道具や柔軟な物体をロボット身体が扱う際に起こる(\secref{subsec:hand-dynamics}, \secref{subsec:cloth-manipulation}で扱う).
  身体と環境の間のモデル化困難性は, 身体が床や障害物等の環境とインタラクションする際に起こる(\secref{subsec:stable-fetch}, \secref{subsec:balance-control}で扱う).
  本研究では直接は扱わないが, 物体/道具と環境の間にも摩擦や作用反作用についてのモデル化困難性は起こりやすい.

  \subsubsection{Classification of Temporal Model Changes}

  最後に, 本研究で扱う逐次的モデル変化を分類する(\figref{figure:classification}の右図).
  STMとCTM, また, ロボットの行動, 身体, 物体/道具, 環境によってそれらを分類することができる.
  ここで, 逐次的なモデル変化は, 予測モデルがその変化する運動や感覚を予測するがゆえに検知できるものである.
  そのため, CTMは行動, 身体, 物体/道具, 環境のそれぞれの逐次的モデル変化を検知することができるが, STMは行動を予測しないため, 行動については逐次的モデル変化を考慮できない.
  行動については, ある同一の状態に対する行動の変化, つまり動作スタイルの変化が挙げられる(\secref{subsec:imitation-learning}で扱う).
  身体については, 経年劣化による身体変化や直接モデルに含まれない部分の身体変化, ロボットが着ている, 履いている物の変化等が挙げられる(CTMにつては\secref{subsec:visual-feedback}, STMについては\secref{subsec:hand-dynamics}と\secref{subsec:balance-control}で扱う).
  物体/道具については, 把持物体の変化や扱う道具の変化, 操作する布の変化等が挙げられる(\secref{subsec:hand-dynamics}と\secref{subsec:cloth-manipulation}で扱う).
  環境については, 床素材の変化や周囲の障害物変化, 壁の摩擦変化等が挙げられる(\secref{subsec:stable-fetch}で扱う).

  これらから, 本研究のDPMPBは, その構造(STMまたはCTM), 行動, 身体, 物体/道具, 環境におけるどれとどれの間のモデル化困難性を扱うのか, そしてどの逐次的モデル変化を扱うのかで, 様々な組み合わせ, 様々なタスクを実現することができる.
  これは具体的には, ネットワーク出力に$\bm{u}$を含むか, $\bm{s}$と$\bm{u}$は何を選ぶか, $\bm{s}$と$\bm{u}$の収集時に何が変化するか, というパラメータの変化に相当する.
  本研究では, 予測モデルにおいてこれらパラメータを変化させるのみで, これまでの分類から考えられるタスク例を網羅的に実現可能であることを示す.
}%

\subsection{Network Structure} \label{subsec:network-structure}
\switchlanguage%
{%
  The network structure of DPMPB proposed in this study (\figref{figure:network-structure}) is shown in the following equations,
  \begin{align}
    \bm{y}_{t+1} &= \bm{h}(\bm{x}_{t}, \bm{p}) \label{eq:dpmpb}\\
    \bm{x}_{t} &= \begin{pmatrix}\bm{s}^{T}_{t}, \bm{u}^{T}_{t}\end{pmatrix}^{T}\nonumber\\
    \bm{y}_{t+1} &= \left\{ \begin{array}{ll} \bm{s}_{t+1} & (type: \textrm{STM}) \\ \begin{pmatrix}\bm{s}^{T}_{t+1}, \bm{u}^{T}_{t+1}\end{pmatrix}^{T} & (type: \textrm{CTM}) \end{array} \right.\nonumber
  \end{align}
  where $t$ is the current time step, $\bm{s}$ is the sensor state of the robot body, target objects, etc., $\bm{u}$ is the control input representing the motion, $\bm{x}$ is the network input, $\bm{y}$ is the network output, $\bm{p}$ is parametric bias (PB), and $\bm{h}$ is the predictive model containing the network weight $\bm{W}$.
  The information contained in $\bm{y}$ differs depending on whether the predictive model type is STM or CTM.
  STM outputs only $\bm{s}$, while CTM outputs $\bm{u}$ in addition.
  In the case of STM, $\bm{u}$ is optimized to make $\bm{s}$ closer to the target state, while in the case of CTM, the next control input can be calculated directly from the current state.
  Note that the network structure of STM or CTM can be automatically determined from the collected data.
  $\bm{p}$ is a low-dimensional input variable that can embed implicit differences in dynamics by collecting data while changing the physical state of the robot, target objects, tools, and environment.

  Here, we briefly describe the applications of this network structure.
  The network basically contains $\bm{s}$, $\bm{u}$, $\bm{p}$, and $\bm{W}$ as values.
  The only operations possible here are to compute these values by forward propagation, or to update them from the loss function by back propagation and gradient descent methods.
  In \figref{figure:network-structure}, the former operation is used for Simulator to update the current $\bm{s}$ (not directly handled in this study), and for Anomaly Detector to take prediction errors for $\bm{s}$.
  The latter operation is used for Network Trainer to update $\bm{W}$ and $\bm{p}$ simultaneously, for Online PB Updater to update only $\bm{p}$, and for Controller to update only $\bm{u}$.

  The basic network structure is described below.
  Our DPMPB is a 10-layer recurrent neural network consisting of four FC layers (fully-connected layers), two LSTM layers (long short-term memory layers \cite{hochreiter1997lstm}), and four FC layers.
  The activation function is hyperbolic tangent and the update rule is Adam \cite{kingma2015adam}.
  In addition, all values of $\bm{s}$ and $\bm{u}$ are normalized and used as the network input and output.
  The control frequency of $\equref{eq:dpmpb}$ is basically 5 Hz (1 Hz only for \secref{subsec:visual-feedback}).
  The dimension of $\bm{p}$ should be slightly smaller than the expected changes in the body state, because too small a dimensionality will not represent the change in dynamics properly, and too large a dimensionality will make self-organization of $\bm{p}$ difficult.
  Regardless of the dimensionality of the changes in dynamics, they will be compressed to the point where they can be represented by the set dimension of PB.
}%
{%
  本研究で提案するDPMPBのネットワーク構造(\figref{figure:network-structure})を以下に式で示す.
  \begin{align}
    \bm{y}_{t+1} &= \bm{h}(\bm{x}_{t}, \bm{p}) \label{eq:dpmpb}\\
    \bm{x}_{t} &= \begin{pmatrix}\bm{s}^{T}_{t}, \bm{u}^{T}_{t}\end{pmatrix}^{T}\\
      \bm{y}_{t+1} &= \left\{ \begin{array}{ll} \bm{s}_{t+1} & (type: \textrm{STM}) \\ \begin{pmatrix}\bm{s}^{T}_{t+1}, \bm{u}^{T}_{t+1}\end{pmatrix}^{T} & (type: \textrm{CTM}) \end{array} \right.
  \end{align}
  ここで, $t$は現在のタイムステップ, $\bm{s}$はロボット身体や対象物体等に関する感覚状態, $\bm{u}$は運動を表す制御入力, $\bm{x}$はネットワーク入力, $\bm{y}$はネットワーク出力, $\bm{p}$はParametric Bias (PB), $\bm{h}$はネットワーク重み$\bm{W}$を含む予測モデルを表す.
  予測モデルのタイプがSTMかCTMかによって$\bm{y}$が含む情報が異なる.
  STMは$\bm{s}$のみ出力するのに対して, CTMは加えて$\bm{u}$も出力する.
  STMの場合は$\bm{s}$を指令状態に近づけるように$\bm{u}$を最適化するのに対して, CTMの場合は現在状態から直接次の制御入力を計算することができる.
  なお, このSTMまたはCTMのネットワーク構造は, データから自動で判断することができる.
  $\bm{p}$は暗黙的なダイナミクスの違いを埋め込むことができる低次元の入力変数であり, ロボットの身体状態や道具, 環境等を変化させながらデータを取得することで, これらの情報が$\bm{p}$の中に暗黙的に自己組織化される.

  ここで, 本ネットワーク構造の応用方法について簡単に述べる.
  本ネットワークは基本的に$\bm{s}$, $\bm{u}$, $\bm{p}$, $\bm{W}$の値を入出力や内部に含む.
  ここで可能な操作は, これらの値を順伝播により計算すること, または損失関数から誤差逆伝播と勾配法により更新することのみである.
  \figref{figure:network-structure}において, 後者の操作によって, $\bm{W}$と$\bm{p}$を同時に更新することをNetwork Trainer, $\bm{p}$のみを更新ことをOnline PB Updater, $\bm{u}$のみ更新することをControllerと呼ぶ.
  また, 前者の操作によって$\bm{s}$のみ順伝播することをSimulator (本研究では直接は扱わない), その$\bm{s}$に対して予測誤差を取ることをAnomaly Detectorと呼ぶ.

  基本的なネットワーク構造について述べておく.
  本研究のDPMPBは全て10層であり, 順に4つのFC層(fully-connected layer), 2層のLSTM層(long short-term memory \cite{hochreiter1997lstm}), 4層のFC層からなる再帰的ネットワークである.
  活性化関数はHyperbolic Tangent, 更新則はAdam \cite{kingma2015adam}としている.
  また, $\bm{s}$と$\bm{u}$の全ての値に正規化を施したうえで, ネットワーク入出力としている.
  LSTMを使う理由だが, 大抵の場合, $\bm{s}$では表現できないような動作系列が影響する要素がロボット状態に含まれるためである.
  \equref{eq:dpmpb}の周期は基本的に5 Hzである(\secref{subsec:visual-feedback}のみ1 Hzとしている).
  $\bm{p}$の次元数は小さ過ぎるとdynamicsの変化を適切に表せなくなり, 大きすぎると自己組織化が難しくなるため, 想定される身体状態変化よりも少し小さく設定することが望ましい.
  dynamics変化の次元の大きさに関わらず, それらはPBの次元で表現できるところまで圧縮されることになる.
}%

\subsection{Data Collection and Training of DPMPB} \label{subsec:network-training}
\switchlanguage%
{%
  First, we collect the time series data of $\bm{s}$ and $\bm{u}$.
  For this purpose, we mainly use two types of data collection methods: (1) random motion and (2) teaching motion.
  Random motion is a motion in which the minimum and maximum values of the control input are determined and the body is moved randomly within these values.
  Teaching motion is a motion in which a human moves the robot body using a VR device, GUI, gaming controller, etc.
  While random motion allows the robot to explore space widely and evenly, teaching motion is effective for tasks that are difficult to succeed with random motion.
  Since CTM requires that the next control input can be computed from the current state, STM is most likely used when handling the data collected by (1), but the network structure will be determined automatically.

  Using the obtained time series data $D$ of $\bm{s}$ and $\bm{u}$, DPMPB is trained.
  In this process, by collecting data while changing the states of the body, target objects, tools, and environment, this information can be implicitly embedded into the space of PB.
  In order to allow the transition of each time series data with different dynamics to be represented by a single model, the differences in dynamics are self-organized in a low-dimensional space of $\bm{p}$.
  The data collected for a given identical state $k$ is represented as $D_{k}=\{(\bm{s}_{1}, \bm{u}_{1}), (\bm{s}_{2}, \bm{u}_{2}), \cdots, (\bm{s}_{T_{k}}, \bm{u}_{T_{k}})\}$ ($1 \leq k \leq K$, where $K$ is the total number of states, and $T_{k}$ is the number of time steps for the trial in the state $k$), and the data used for training $D_{train}=\{(D_{1}, \bm{p}_{1}), (D_{2}, \bm{p}_{2}), \cdots, (D_{K}, \bm{p}_{K})\}$ is obtained.
  Here, $\bm{p}_{k}$ is PB that represents the dynamics with respect to the state $k$.
  Since $\bm{p}_{k}$ is a variable, the initial value can be anything, just like the bias term in a neural network, and $\bm{p}_{k}$ does not require a specific value for training.
  PB is a common variable for a particular state but is another variable for another state.
  Using this $D_{train}$, we train DPMPB with the number of network expansions as $N^{train}_{step}$, the number of batches as $N^{train}_{batch}$, and the number of epochs as $N^{train}_{epoch}$.
  In the usual training, only the network weight $\bm{W}$ is updated, but here $\bm{W}$ and $\bm{p}_{k}$ for each state are updated simultaneously.
  Note that the mean squared error for the network output $\bm{y}$ is used as the loss function in the learning process, and $\bm{p}_{k}$ is optimized with an initial value of $\bm{0}$ for all.
  In this way, the differences in the dynamics of each body state are embedded in $\bm{p}_{k}$.

  Here, we also describe the automatic determination of the network structure of STM or CTM.
  First, we train the network as $\bm{y}_{t+1}=\begin{pmatrix}\bm{s}^{T}_{t+1}, \bm{u}^{T}_{t+1}\end{pmatrix}^{T}$.
  In this case, we calculate the loss $L_{n}$ separately for each value of $\bm{y}_{n}$ ($1 \leq n \leq N_{sensor}$, where $N_{sensor}$ represents the number of sensors).
  We set a threshold $L_{thre}$ and adopt the value with $L_{n}$ smaller than $L_{thre}$ as $\bm{y}$.
  If $\bm{y}$ includes the value of $\bm{u}$, it is CTM, and if not, it is STM.
  Thus, by removing the values that are difficult to infer from the network output, the network structure is defined and the possible tasks in the network are changed.
  The network is then re-trained with the adopted $\bm{y}$.
}%
{%
  まず, $\bm{s}$と$\bm{u}$の時系列データを収集する.
  これには(1)ランダム動作と(2)教示動作の2種類のデータ収集方法を主に利用した.
  ランダム動作はランダムに動かす制御入力とその最小値最大値を決定し, その中でランダムに身体を動かす動作である.
  教示動作は, 人間がVRデバイスやGUI, ゲーミングコントローラ等を使い, それに沿ってロボットの身体を動かす動作である.
  ランダム動作は広く万遍なく空間を探索できるのに対して, 教示動作はランダム動作では成功しにくいタスクに対して有効である.
  CTMは現在状態から次の制御入力が計算可能である必要があるため, (1)のデータを用いる場合はSTMになる可能性が高いが, これらは自動で判断することもできる.

  得られた$\bm{s}$と$\bm{u}$の時系列データ$D$を使いDPMPBを学習させる.
  この際, 身体や道具, 環境の状態を変化させながらデータを収集することで, これらの情報を暗黙的にPBに埋め込むことができる.
  異なるダイナミクスを持つそれぞれの時系列データ遷移を一つのモデルで表現できるように, そのダイナミクスの違いを低次元の$\bm{p}$の空間に形作る.
  ある同一の状態$k$について収集されたデータを$D_{k}=\{(\bm{s}_{1}, \bm{u}_{1}), (\bm{s}_{2}, \bm{u}_{2}), \cdots, (\bm{s}_{T_{k}}, \bm{u}_{T_{k}})\}$ ($1 \leq k \leq K$, $K$は全試行回数, $T_{k}$はその身体状態$k$における試行の動作ステップ数)として, 学習に用いるデータ$D_{train}=\{(D_{1}, \bm{p}_{1}), (D_{2}, \bm{p}_{2}), \cdots, (D_{K}, \bm{p}_{K})\}$を得る.
  ここで, $\bm{p}_{k}$はそのデータ収集時の状態$k$に関するダイナミクスを表現するPBであり, その状態については共通の変数, 別の状態については別の変数となる.
  あくまで変数であるため, ニューラルネットワークにおけるbias項と同じく初期値は何でもよく, 具体的な値が必要なものではない.
  この$D_{train}$を使い, ネットワークの展開数を$N^{train}_{step}$, バッチ数を$N^{train}_{batch}$, エポック数を$N^{train}_{epoch}$としてDPMPBを学習させる.
  通常の学習ではネットワークの重み$\bm{W}$のみが更新されるが, ここでは$\bm{W}$と各状態に関する$\bm{p}_{k}$が同時に更新される.
  なお, 学習の際は損失関数としてネットワーク出力$\bm{y}$に対する平均二乗誤差を使い, $\bm{p}_{k}$は全て$\bm{0}$を初期値として最適化される.
  これにより, $\bm{p}_{k}$にそれぞれの身体状態におけるダイナミクスの違いが埋め込まれることになる.

  ここで, STMまたはCTMのネットワーク構造の自動判断についても述べる.
  始めに, $\bm{y}_{t+1}=\begin{pmatrix}\bm{s}^{T}_{t+1}, \bm{u}^{T}_{t+1}\end{pmatrix}^{T}$としてネットワークを学習させる.
  このとき, $\bm{y}_{n}$ ($1 \leq n \leq N_{sensor}$, $N_{sensor}$はセンサ数を表す)のそれぞれの値に対して別々に損失$L_{n}$を計算しておく.
  ある閾値$L_{thre}$を設定し, この値よりも小さな$L_{n}$を持つ値を$\bm{y}$として採用する.
  この$\bm{y}$に$\bm{u}$の値が含まれていればCTMであり, 含まれていなければSTMである.
  そもそも推論の難しい値をネットワーク出力から削除することで, ネットワーク構造が規定され, そのネットワークで可能なタスクが変化する.
  その後, 定まった$\bm{y}$で再度学習し直す.
}%

\subsection{Online Update of Parametric Bias} \label{subsec:online-update}
\switchlanguage%
{%
  $\bm{p}_{k}$ computed at the time of training represents the dynamics corresponding to each data $D_{k}$ and does not represent the current dynamics.
  Therefore, we update PB online using the data $D$ obtained from the current state of the body, target objects, tools, and environment.
  If the network weight $\bm{W}$ is updated, DPMPB may overfit to the data, but if only the low-dimensional PB $\bm{p}$ is updated, overfitting is less likely to occur.
  This online learning allows us to constantly recognize the current robot state and to obtain a model that adapts to it.

  Let the number of data obtained be $N^{online}_{data}$, and start online learning when the number of data exceeds the threshold $N^{online}_{thre}$.
  Whenever new data is received, PB is updated with the number of batches as $N^{online}_{batch}$, the number of epochs as $N^{online}_{epoch}$, and the update rule as MomentumSGD.
  Data exceeding the maximum number of data $N^{online}_{max}$ are deleted from the oldest ones.
  The smaller $N^{online}_{max}$ is, the faster the model may adapt to the current state, but the learning may become unstable due to the decrease in the number of data.
}%
{%
  Training時に計算された$\bm{p}_{k}$はそれぞれのデータ$D_{k}$に対応するダイナミクスを表現しており, 現在のダイナミクスを表現できてはいない.
  そこで, 現在の身体や道具, 環境の状態において得られたデータ$D$を使い, オンラインでPBを更新する.
  ネットワークの重み$\bm{W}$を更新してしまうとDPMPBがそのデータに過学習してしまう可能性があるが, 低次元のPB $\bm{p}$のみを更新するのであれば過学習は起こらない.
  このオンライン学習により, 常に現在の身体状態を認識し, これに適応したモデルを得ることができる.

  得られたデータ数を$N^{online}_{data}$として, データ数が$N^{online}_{thre}$を超えたところからオンライン学習を始める.
  新しいデータが入るたびにバッチ数を$N^{online}_{batch}$, エポック数を$N^{online}_{epoch}$, 更新則をMomentumSGDとして学習を行う.
  $N^{online}_{max}$を超えたデータは古いものから削除していく.
  $N^{online}_{max}$が小さい方がより素早く現在状態に適応できるが,データ数が減るため学習が不安定になる可能性がある.
}%

\subsection{Control Using DPMPB} \label{subsec:control}
\switchlanguage%
{%
  We describe a control method using DPMPB.
  This is very different depending on whether the network structure is CTM or STM.
  In the case of CTM, it is very simple; since $\bm{u}$ is in the output, we only need to calculate $\bm{u}_{t}$ from $\bm{s}_{t-1}$ and $\bm{u}_{t-1}$ of the previous time step through the forward propagation of the network, and send it to the robot.

  On the other hand, in the case of STM, since $\bm{u}$ is not in the output, we need to optimize $\bm{u}$ from a loss function.
  This is a kind of learning-based model predictive control.
  First, we give the initial value of time-series control input $\bm{u}^{init}_{seq}$ for $\bm{u}_{seq}=\bm{u}_{t:t+N^{control}_{step}-1}$ ($\bm{u}_{t_1:t_2}$ is $\bm{u}$ in [$t_1$, $t_2$]).
  $N^{control}_{step}$ is the number of DPMPB network expansions representing control horizon (the expansion operation will be explained later).
  Let $\bm{u}^{opt}_{seq}$ be $\bm{u}_{seq}$ to be optimized, and repeat the following calculation at time $t$ to obtain the optimal $\bm{u}^{opt}_{t}$,
  \begin{align}
    \bm{s}^{pred}_{seq} &= \bm{h}_{expand}(\bm{s}_{t}, \bm{u}^{opt}_{seq}, \bm{p})\\
    L &= h_{loss}(\bm{s}^{pred}_{seq}, \bm{u}^{opt}_{seq}) \label{eq:control-loss}\\
    \bm{u}^{opt}_{seq} &\gets \bm{u}^{opt}_{seq} - \gamma\partial{L}/\partial{\bm{u}^{opt}_{seq}} \label{eq:control-opt}
  \end{align}
  where $\bm{s}^{pred}_{seq}$ is the predicted $\bm{s}_{t+1:t+N^{control}_{step}}$, $\bm{h}_{expand}$ is the function of $\bm{h}$ expanded $N^{control}_{step}$ times, $h_{loss}$ is the loss function, and $\gamma$ is the learning rate.
  Note that the network expansion in $\bm{h}_{expand}$ is a function that represents the operation of inputting $\bm{u}^{opt}$ from $\bm{s}_{t}$ and predicting $\bm{s}^{pred}$ in $N^{control}_{step}$ steps.
  In other words, the future $\bm{s}$ is predicted from $\bm{s}_{t}$ by $\bm{u}^{opt}_{seq}$, and $\bm{u}^{opt}_{seq}$ is optimized by backpropagation and gradient descent to minimize the loss function.

  Here, $\bm{u}^{init}_{seq}$ is set to $\bm{u}^{prev}_{\{t+1, \cdots, t+N^{control}_{step}-1, t+N^{control}_{step}-1\}}$, by using $\bm{u}^{prev}_{t:t+N^{control}_{step}-1}$ ($\bm{u}_{seq}$ optimized in the previous step), and by shifting the time step of the value by one and duplicating the last term.
  Faster convergence can be obtained by using the previous optimization results.
  For $\gamma$, we set the maximum learning rate $\gamma_{max}$, prepare $N^{control}_{batch}$ of $\gamma$ by dividing $[0, \gamma_{max}]$ exponentially, run \equref{eq:control-opt} on each $\gamma$, calculate the loss by \equref{eq:control-loss}, and select $\bm{u}^{opt}_{seq}$ with the smallest loss.
  This procedure is repeated $N^{control}_{epoch}$ times.
  Faster convergence can be obtained by trying various $\gamma$ and always choosing the best learning rate.

  There are various possible forms of the loss function $h_{loss}$.
  For each value of $\bm{s}$ and $\bm{u}$, the main possible forms are minimization of the value, maximization of the value, minimization of the error with a certain target value, and minimization of the change from the value of the previous time step.
  These are set appropriately for each experiment.
}%
{%
  DPMPBを使った制御手法について述べる.
  これは, ネットワーク構造がCTMかSTMかによって大きく異なる.
  CTMの場合は非常に単純で, $\bm{u}$が出力にあるため, 前時刻の$\bm{s}_{t-1}$と$\bm{u}_{t-1}$からネットワークの順伝播を通して$\bm{u}_{t}$を計算, これをロボットに送るのみで良い.

  一方, STMの場合は$\bm{u}$が出力にないため, 損失関数から$\bm{u}$を最適化していく必要がある.
  これは学習型のモデル予測制御の一種でもある.
  まず, 時系列制御入力$\bm{u}_{seq}=\bm{u}_{t:t+N^{control}_{step}-1}$の初期値$\bm{u}^{init}_{seq}$を与える.
  ここで, $N^{control}_{step}$は制御ホライゾンを表すDPMPBのネットワークの展開数である(展開の操作については後に説明する).
  最適化する$\bm{u}_{seq}$を$\bm{u}^{opt}_{seq}$とおき, 時刻$t$において以下の計算を繰り返すことで最適な$\bm{u}^{opt}_{t}$を得る.
  \begin{align}
    \bm{s}^{pred}_{seq} &= \bm{h}_{expand}(\bm{s}_{t}, \bm{u}^{opt}_{seq}, \bm{p})\\
    L &= h_{loss}(\bm{s}^{pred}_{seq}, \bm{u}^{opt}_{seq}) \label{eq:control-loss}\\
    \bm{u}^{opt}_{seq} &\gets \bm{u}^{opt}_{seq} - \gamma\partial{L}/\partial{\bm{u}^{opt}_{seq}} \label{eq:control-opt}
  \end{align}
  ここで, $\bm{s}^{pred}_{seq}$は予測された$\bm{s}_{t+1:t+N^{control}_{step}}$, $\bm{h}_{expand}$は$\bm{h}$を$N^{control}_{step}$回展開した関数, $h_{loss}$は損失関数, $\gamma$は学習率を表す.
  なお, $\bm{h}_{expand}$におけるネットワーク展開は, $\bm{s}_{t}$から$\bm{u}^{opt}$を入力し$\bm{s}^{pred}$を予測することを$N^{control}_{step}$ステップ行うという操作を一つの関数で表現したものである.
  つまり, $\bm{s}_{t}$から$\bm{u}^{opt}_{seq}$により将来の$\bm{s}$を予測し, これに対して設定した損失関数を最小化するように, $\bm{u}^{opt}_{seq}$を誤差逆伝播法と勾配降下法により最適化する.

  このとき$\bm{u}^{init}_{seq}$は, 前ステップで最適化された$\bm{u}_{seq}$である$\bm{u}^{prev}_{t:t+N^{control}_{step}-1}$を使い, 時間を一つシフトし最後の項を複製した$\bm{u}^{prev}_{\{t+1, \cdots, t+N^{control}_{step}-1, t+N^{control}_{step}-1\}}$とする.
  前回の最適化結果を利用することでより速い収束が得られる.
  また, $\gamma$については学習率の最大値$\gamma_{max}$を設定し, $[0, \gamma_{max}]$を指数関数的に分割した$N^{control}_{batch}$個の$\gamma$を用意し, それぞれの$\gamma$で\equref{eq:control-opt}を実行した後, \equref{eq:control-loss}で損失を計算し最も損失の小さい$\bm{u}^{opt}_{seq}$を選択することを$N^{control}_{epoch}$回繰り返す.
  様々な$\gamma$を試して最良の学習率を常に選ぶことで, より速い収束が得られる.

  損失関数$h_{loss}$には様々な形式が考えられる.
  $\bm{s}$と$\bm{u}$のそれぞれの値について, 最小化, 最大化, ある指令値との誤差最小化, 前ステップの値からの変化最小化等が主に考えられる.
  これについては, その実験ごとに適切に設定している.
}%

\subsection{Anomaly Detection Using DPMPB}\label{subsec:anomaly-detection}
\switchlanguage%
{%

  Anomaly detection can be performed from the prediction error of DPMPB.
  This is executed for the magnitude of the error between the current value $\bm{s}_{t}$ and the predicted value $\bm{s}^{pred}_{t}$ for the output of the network.
  First, for the data $D$ collected in the normal state without any anomaly, we collect the measured value $\bm{s}^{data}_{t}$ and the predicted value $\bm{y}_{t}=\bm{h}(\bm{x}_{t-1}, \bm{p})$ calculated from the previous time step.
  For this data, the mean $\bm{\mu}$ and variance $\bm{\Sigma}$ of the error $\bm{s}^{data}_{t}-\bm{s}^{pred}_{t}$ is calculated.
  For anomaly detection, $\bm{e}_{t}=\bm{s}^{pred}_{t}-\bm{s}_{t}$ (which is the difference between the current state value $\bm{s}_{t}$ and the estimated value $\bm{s}^{pred}_{t}$) is always obtained, and the Mahalanobis distance $d=\sqrt{(\bm{e}_{t}-\bm{\mu})^{T}\bm{\Sigma}^{-1}(\bm{e}_{t}-\bm{\mu})}$ is calculated for them.
  When $d$ exceeds a threshold value, an anomaly has been detected.
  For this threshold, the variance of $d$ can be calculated for the data used to calculate $\bm{\mu}$ and $\bm{\Sigma}$, and 3$\sigma$ value can be used.
  Anomaly can be calculated for each sensor or for all sensors at once.
}%
{%
  DPMPBの予測誤差から異常検知を行うことができる.
  これはネットワークの状態出力に関する現在値$\bm{s}_{t}$と予測値$\bm{s}^{pred}_{t}$の誤差の大きさに対して行う.
  まず, 異常のない, 正常な状態において収集したデータ$D$について, 実測値$\bm{s}^{data}_{t}$と前ステップから$\bm{y}_{t}=\bm{h}(\bm{x}_{t-1}, \bm{p})$により計算された予測値$\bm{s}^{pred}_{t}$を収集しておく.
  このデータに対して, 誤差$\bm{s}^{data}_{t}-\bm{s}^{pred}_{t}$の平均$\bm{\mu}$と分散$\bm{\Sigma}$を計算しておく.
  実際に異常検知をする際には, $\bm{s}$に関する現在値$\bm{s}_{t}$と推定値$\bm{s}^{pred}_{t}$の差$\bm{e}_{t}=\bm{s}^{pred}_{t}-\bm{s}_{t}$を常に取得し, これらに対して以下のマハラノビス距離$d$を計算する.
  \begin{align}
    d = \sqrt{(\bm{e}_{t}-\bm{\mu})^{T}\bm{\Sigma}^{-1}(\bm{e}_{t}-\bm{\mu})}
  \end{align}
  この$d$が設定した閾値を超えた時, 異常が検知されたこととする.
  この閾値であるが, $\bm{\mu}$と$\bm{\Sigma}$を計算する際に用いられたデータに対して$d$の分散を計算しておき, その3$\sigma$区間等を用いることも可能である.
  なお, この異常度は, センサ一つずつに対して計算しても良いし, 全てのセンサに対して一度に計算しても良い.
}%

\begin{figure}[t]
  \centering
  \includegraphics[width=0.99\columnwidth]{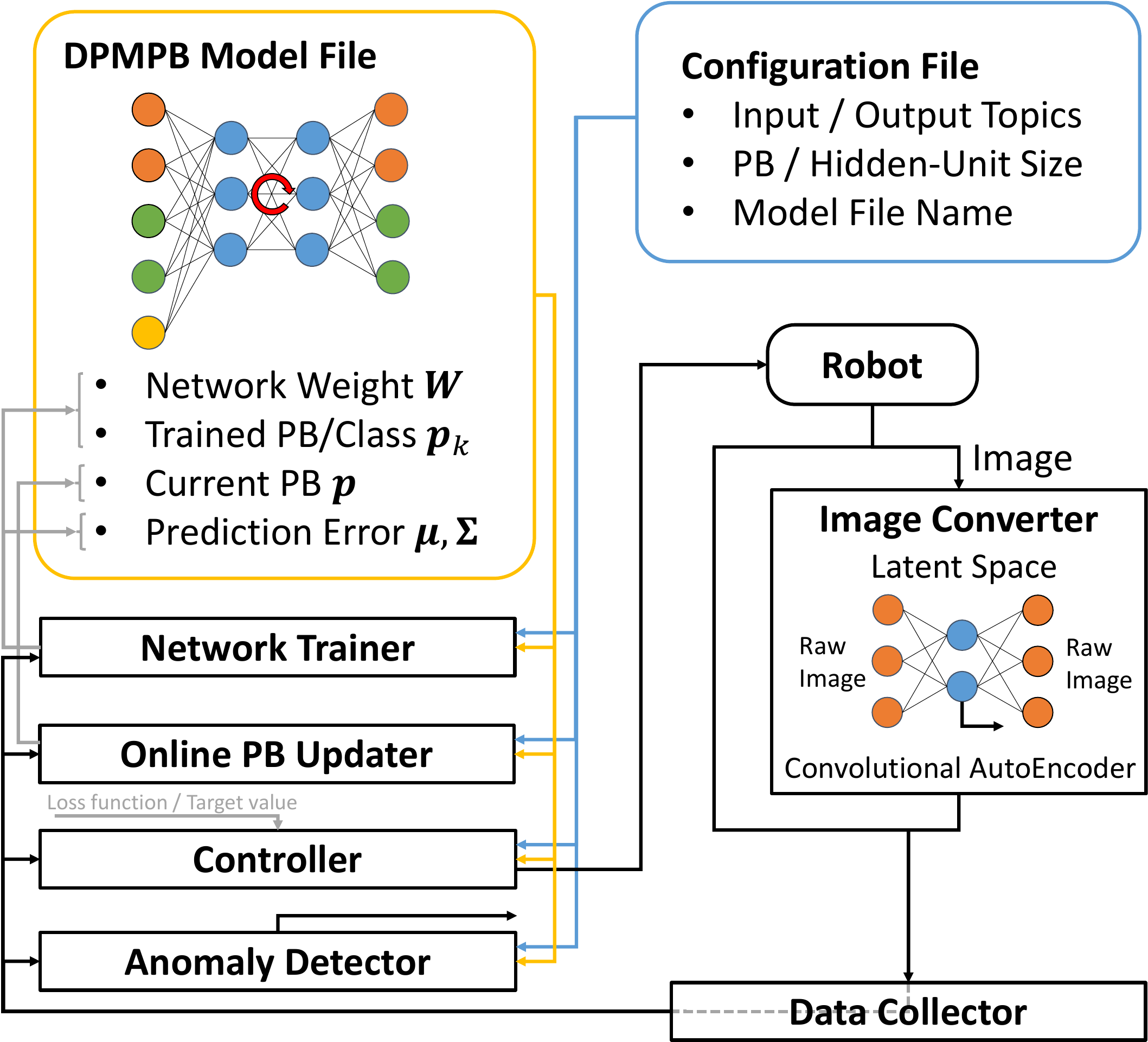}
  \vspace{-3.0ex}
  \caption{The detailed implementation of deep predictive model with parametric bias (DPMPB). The model file includes the information of network weight $\bm{W}$, trained parametric bias (PB) $\bm{p}_{k}$ and class, current PB $\bm{p}$, and the average $\mu$ and variance $\Sigma$ of the prediction error calculated at training. The configuration file includes input/output topics of Robot Operating System (ROS), the dimension of PB and hidden unit of DPMPB, and the model file name. The image is compressed by AutoEncoder and Data Collector gathers and sends all data to each component of Network Trainer, Online PB Updater, Controller, and Anomaly Detector. Network Trainer updates $\bm{W}$ and $\bm{p}_{k}$, and Online PB Updater updates $\bm{p}$.}
  \label{figure:network-detailed}
  \vspace{-3.0ex}
\end{figure}

\subsection{Detailed Implementation}\label{subsec:detailed-implementation}
\switchlanguage%
{%
  We describe the implementation of DPMPB in more detail (\figref{figure:network-detailed}).
  First of all, we use ROS (Robot Operating System) for all sensor communication and Chainer as the deep learning framework in this study.
  The configuration file describes which ROS topic is used for the input and output of DPMPB, the number of PB and hidden layer units in the network, and the model file name.
  The model file of the network contains the network weight $\bm{W}$, the class name for each state obtained during training and the corresponding PB $\bm{p}_{k}$, the current PB $\bm{p}$, the mean $\bm{\mu}$ and variance $\bm{\Sigma}$ of the prediction error of the network.
  Note that by storing the class names, explicit object and environment recognition can be performed from $\bm{p}_{k}$, which is closest to the current $\bm{p}$.
  Although the sensory and motion data of the robot are collected as is, images are processed in a special way.
  Since the data size of an image is too large as it is, we train Convolutional AutoEncoder in advance and use the value converted into the latent space as a ROS Topic.
  Data Collector in \figref{figure:network-detailed} is drawn differently from \figref{figure:network-structure} for convenience, but it summarizes the obtained data for Network Trainer, Online PB Updater, Controller, and Anomaly Detector.
  For Network Trainer, Online PB Updater, and Controller, the number of batches, the number of epochs, and the learning rate need to be given as parameters, and for Controller, the form of the loss function and the target value need to be given in addition.
  Network Trainer stores $\bm{W}$ and $\bm{p}_{k}$, and Online PB Updater updates only $\bm{p}$.
}%
{%
  より具体的な実装について述べる.
  まず, 本研究の全体のセンサのやり取りにはROS (Robot Operating System)を用いている.
  設定ファイルには, DPMPBの入力と出力にどのROS Topicを用いるか, また, ネットワークにおけるPBや隠れ層のユニット数, モデルファイル名が記述されている.
  ネットワークのモデルファイルにはネットワーク重み$\bm{W}$, 学習時に得られるそれぞれの状態に関するクラス名と対応するPB $\bm{p}_{k}$, 現在のPB $\bm{p}$, ネットワークの予測誤差の平均$\bm{\mu}$と分散$\bm{\Sigma}$が含まれている.
  なお, クラス名を保存しておくことで, 現在の$\bm{p}$に最も近い$\bm{p}_{k}$から, 明示的な物体認識や環境認識が可能になる.
  ロボットから得られた感覚と運動のデータを収集していくが, このとき画像については特殊な処理を挟む.
  画像はそのままではデータサイズが大きいため, 事前にConvolutional AutoEncoderを学習しておき, これを使って潜在空間に変換した値をROS Topicとして出力しておく.
  Data Collectorは便宜上\figref{figure:network-structure}と多少書き方が異なるが, 得られたデータを使い, Network Trainer, Online PB Updater, Controller, Anomaly Detectorが働いている.
  Network Trainer, Online PB Updater, Controllerには, パラメータとしてバッチ数, エポック数, 学習率を与える必要があり, Controllerのみ追加で損失関数の形と指令値を与える必要がある.
  Network Trainerは$\bm{W}$と$\bm{p}_{k}$を保存し, Online PB Updaterは$\bm{p}$のみ更新する.
}%

\section{Experiments} \label{sec:experiment}
\switchlanguage%
{%
  In this section, we describe several examples of robot tasks using DPMPB, focusing on what kind of modeling difficulties and temporal model changes are handled, what kind of sensors and actuators are handled, and what kind of loss functions are used in training, online update, and control.
  First, we show the basic experiments of STM and CTM in \secref{subsec:hand-dynamics} and \secref{subsec:visual-feedback}, respectively, to demonstrate their functions.
  Next, in \secref{subsec:stable-fetch}, we show an experiment of variance minimization control by introducing mean-variance representation to the network output as an application of STM, and in \secref{subsec:imitation-learning}, we show an experiment of actively changing the motion style in imitation learning as an application of CTM.
  Finally, \secref{subsec:balance-control} and \secref{subsec:cloth-manipulation} show examples of highly difficult modeling, specifically a balance control experiment considering change in shoes and a dynamic cloth manipulation considering change in cloth material by musculoskeletal humanoids.
  We summarize \cite{kawaharazuka2020dynamics, kawaharazuka2022vservoing, kawaharazuka2021fetch, kawaharazuka2021imitation, kawaharazuka2022balance, kawaharazuka2022cloth} in a new angle from the viewpoint of modeling difficulties and temporal model changes.
  Note that the same symbols are used with different definitions in each experiment.
  Note also that the results of control with the wrong parametric bias are shown for the case where the model at the time of training is different from the current model in the general predictive model, and that these are the results of comparison between DPMPB and the general predictive model.
}%
{%
  ここでは, DPMPBを使った動作例を, どのようなモデル化困難性と逐次的モデル変化を扱っているのか, どのような感覚と運動を扱っているのか, 学習やオンライン学習, 制御の際の損失関数は何であるか等に焦点を当てて述べていく.
  まず, \secref{subsec:hand-dynamics}と\secref{subsec:visual-feedback}においてそれぞれSTMとCTMの基本的な実験を行いその機能を示す.
  次に, \secref{subsec:stable-fetch}ではSTMの応用としてネットワーク出力に平均分散表現を導入し分散最小化制御を行った例, \secref{subsec:imitation-learning}ではCTMの応用として模倣動作における動作スタイルを能動的に変更する例を示す.
  最後に, よりモデル化困難な例として, \secref{subsec:balance-control}と\secref{subsec:cloth-manipulation}において, 全身筋骨格ヒューマノイドMusashiにおける靴変化を考慮したバランス制御と布素材変化を考慮した動的柔軟布操作実験を行う.
  \secref{subsec:hand-dynamics}, \secref{subsec:stable-fetch}, \secref{subsec:imitation-learning}は\cite{kawaharazuka2020dynamics, kawaharazuka2022vservoing, kawaharazuka2021fetch, kawaharazuka2021imitation, kawaharazuka2022balance, kawaharazuka2022cloth}をモデル化困難性と逐次的モデル変化の観点から新しく捉え直してまとめたものである.
  なお, それぞれの実験で同じ文字を異なる定義で使用している点に注意されたい.
  また, Parametric Biasが間違った状態における制御の結果を示しているが, これは, 一般的な予測モデルにおいて, 学習時のモデルと現在のモデルが異なってしまった場合を示しており, 本研究のDPMPBと一般的な予測モデルの比較結果である点にも注意されたい.
}%

\begin{figure}[t]
  \centering
  \includegraphics[width=0.88\columnwidth]{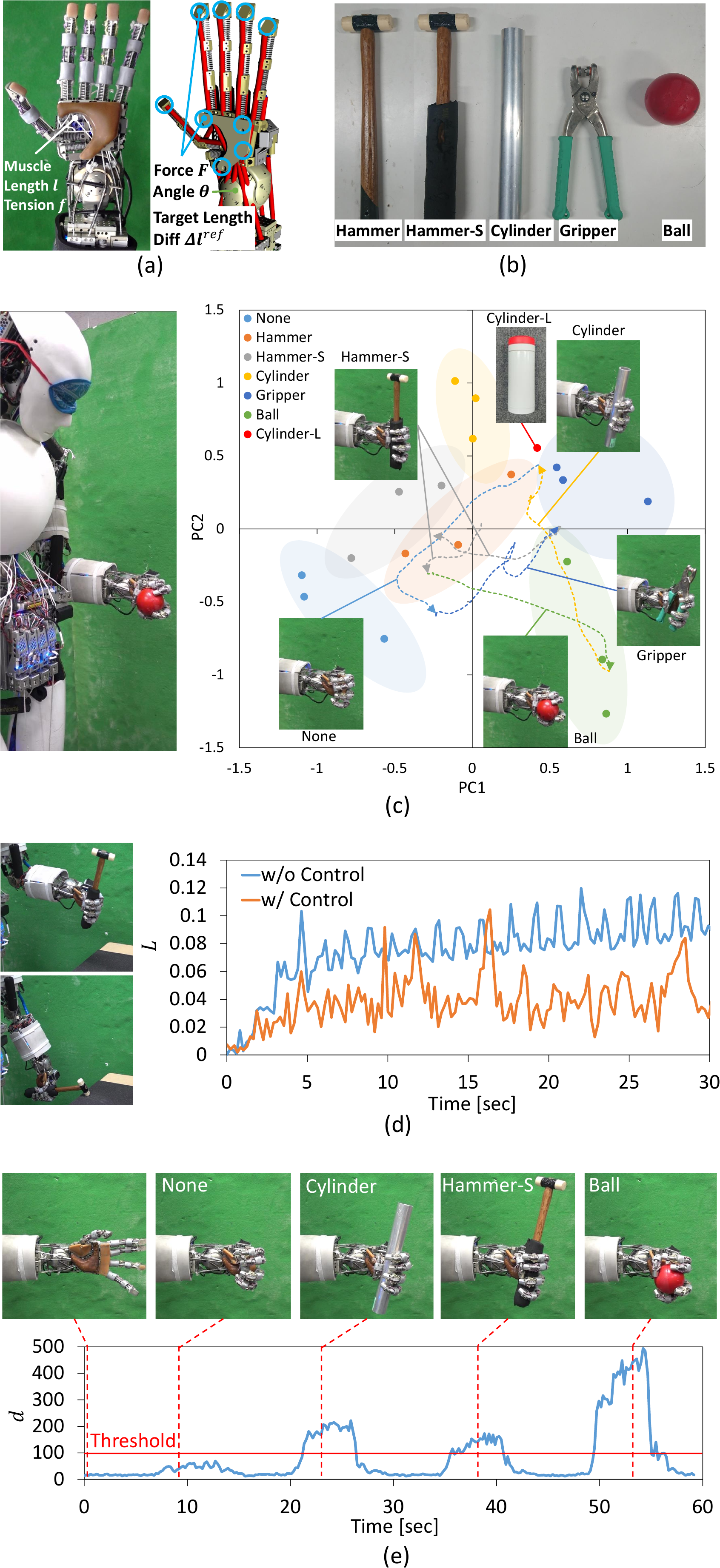}
  \vspace{-2.0ex}
  \caption{Experiment of grasping object recognition and contact control of the flexible hand \cite{kawaharazuka2020dynamics}. (a) shows the sensors and actuators of the flexible musculoskeletal hand, (b) shows the grasped objects and tools, (c) shows the trained parametric bias and its trajectory when conducting online update of PB, (d) shows the transition of $L$ when using or not using a grasping stabilization control, and (e) shows the transition of $d$ for contact detection when grasping various objects.}
  \label{figure:hand-dynamics}
  \vspace{-3.0ex}
\end{figure}

\subsection{Grasping Object Recognition and Contact Control of Flexible Hand} \label{subsec:hand-dynamics}
\switchlanguage%
{%
  In this section, we deal with the sensory-motor model of a musculoskeletal flexible hand \cite{kawaharazuka2019musashi} as a modeling difficulty.
  As a temporal model change, we deal with the change in grasped objects/tools.
  For sensor state and control input, we set $\bm{s}=\{\bm{l}, \bm{f}, \bm{\theta}, \bm{F}\}$ and $\bm{u}=\Delta\bm{l}^{ref}$.
  Note that, as shown in (a) of \figref{figure:hand-dynamics}, $\{\bm{l}, \bm{f}, \Delta\bm{l}^{ref}\}$ is \{muscle length, muscle tension, change in target muscle length\} of the muscles related to the wrist and hand $\in\mathbb{R}^{8}$, $\bm{\theta}$ is the wrist joint angle $\in\mathbb{R}^{2}$, and $\bm{F}$ is the contact sensor value $\in\mathbb{R}^{9}$.
  In (b) of \figref{figure:hand-dynamics}, we show the grasped objects/tools: Hammer, Hammer-S, Cylinder, Gripper, and Ball.
  None refers to the state without grasping any objects/tools.
  This experiment is difficult in that it is necessary to acquire the relationship between five different sensors and actuators in a flexible body, and to recognize the changes in the dynamics of the sensors and actuators and control them accordingly.
  We collected data by randomly changing $\bm{u}$ while changing the grasped objects/tools.
  DPMPB with 8-dimensional PB was trained using 36 datasets with about 500 steps each (about 18000 steps in total).
  Here, $L_{1}=0.093$, $L_{2}=0.184$, $L_{3}=0.212$, $L_{4}=0.036$, and $L_{5}=0.468$, and since $\bm{y}_{5}=\bm{u}$ is removed from the output by setting $L_{thre}=0.3$, the network structure is STM.
  The final loss when training was 0.014.

  (c) of \figref{figure:hand-dynamics} shows the arrangement of the trained PBs $\bm{p}_{k}$.
  Note that PBs are shown after being compressed to two dimensions by Principle Component Analysis (PCA) in all of the following experiments.
  Also, shading in (c) of \figref{figure:hand-dynamics} is manually generated for the set of PBs for better understanding (the same with subsequent experiments).
  It can be seen that the space of PB is self-organized by the dynamics of grasped objects/tools.
  When we collect data for new object Cylinder-L (upper part of (c) in \figref{figure:hand-dynamics}), which is the same shape as Cylinder and has the same radius as Gripper, and update only PB, we can see that PB of Cylinder-L is at about halfway between the PBs of Cylinder and Gripper, and that the space of PB is generalized even for untrained objects/tools.
  In addition, from the trajectory of PB shown as the dotted line in (c), when we perform random motion while changing the grasped objects/tools and simultaneously executing online learning of PB, the current PB $\bm{p}$ gradually approaches the trained PB $\bm{p}_{k}$ for the current grasped object/tool.
  In other words, it is possible to perform recognition of grasped object/tool from its dynamics.
  Here, the average number of data steps used for online update is 500, which indicates that the adaptation takes about 100 seconds.

  In (d) of \figref{figure:hand-dynamics}, we show a grasping stabilization control experiment using DPMPB.
  It is possible to maintain the initial grasping state against external forces by setting $h_{loss}$ in \secref{subsec:control} as follows,
  \begin{align}
    h_{loss}(\bm{s}^{pred}_{seq}) = &||\bm{w}_{1}\otimes(\bm{F}^{pred}_{seq}-\bm{F}^{init}_{seq})||_{2} \nonumber\\
    &+w_{2}||\bm{\theta}^{pred}_{seq}-\bm{\theta}^{init}_{seq}||_{2} + w_{3}||\bm{l}^{pred}_{seq}-\bm{l}^{init}_{seq}||_{2} \label{eq:hand-control-loss}
  \end{align}
  where $\{\bm{F}, \bm{\theta}, \bm{l}\}^{init}_{seq}$ denotes the initial $\{\bm{F}, \bm{\theta}, \bm{l}\}$ when grasping the object/tool, $\{\bm{w}_{1}, w_{2}, w_{3}\}$ denotes the constant weight, $||\cdot||_{2}$ denotes L2 norm, and $\otimes$ denotes element-wise product.
  Considering the anisotropy, where the values of the contact sensor vary only up to 0 in the negative direction, but vary significantly up to the rated value in the positive direction, we set $\bm{w}_{1}[i]=1.0$ when $\bm{F}^{pred}_{seq}[i] \geq \bm{F}^{ref}_{seq}[i])$, otherwise $\bm{w}_{1}[i]=w_4 (w_4 > 1.0)$.
  The graph of (d) shows that when hitting a desk with a hammer, $L$ expressed by $h_{loss}$ gradually increases without this grasping stabilization control, while with it, it is possible to resist large external forces and keep $L$ small.

  In (e) of \figref{figure:hand-dynamics}, we show an application experiment of anomaly detection using DPMPB.
  With the current $\bm{p}$ as $\bm{p}_{k}$ trained at None, grasping an object/tool changes the dynamics of the hand and causes prediction errors in the network.
  This is captured by the anomaly detection of \secref{subsec:anomaly-detection}, which enables us to perform contact detection.
  Note that the farther the grasped object is from None in the space of PB, the larger the anomaly state $d$ is likely to be.
}%
{%
  本研究はモデル化困難性として, 筋骨格柔軟ハンド\cite{makino2018hand}の感覚-運動モデルを扱う.
  また, 逐次的モデル変化として, 把持物体/道具の変化を扱う.
  感覚と運動については, $\bm{s}=\{\bm{l}, \bm{f}, \bm{\theta}, \bm{F}\}$, $\bm{u}=\Delta\bm{l}^{ref}$とした.
  なお, \figref{figure:hand-dynamics}の(a)に示すように, $\{\bm{l}, \bm{f}, \Delta\bm{l}^{ref}\}$は手首と手に関係する筋の\{筋長, 筋張力, 指令筋長変化\} $\in\mathbb{R}^{8}$, $\bm{\theta}$は手首の関節角度センサ値$\in\mathbb{R}^{2}$, $\bm{F}$は接触センサ値$\in\mathbb{R}^{9}$である.
  \figref{figure:hand-dynamics}の(b)には, 用いる把持物体/道具であるHammer, Hammer-S, Cylinder, Gripper, Ballを示す.
  なお, Noneは何も把持していない状態を指す.
  本実験は, 柔軟身体において5種類ものセンサとアクチュエータのモーダルの関係を獲得して制御し, かつそのダイナミクスの変化を認識してそれに応じて制御を行う必要がある点で難しい.
  把持物体/道具を変化させながら, ランダムな$\bm{u}$の変化によりデータを収集する.
  PBを8次元として, 約500ステップのデータを36回, 計約18000ステップのデータを用いて学習している.
  この際, $L_{1}=0.093$, $L_{2}=0.184$, $L_{3}=0.212$, $L_{4}=0.036$, $L_{5}=0.468$となり, $L_{thre}=0.3$と設定することで, $\bm{y}_{5}=\bm{u}$が出力から削除されSTMとなった.
  なお, 最終的な訓練時の推論誤差は0.014であった.

  \figref{figure:hand-dynamics}の(c)には学習されたPB $\bm{p}_{k}$の配置を示す.
  なお, 以降の全ての実験について, PBはPrinciple Component Analysis (PCA)によって2次元に圧縮されたうえで表示されている.
  また, 一部わかりやすいようにPBの集合に対して網掛けを手動で作成している.
  PBが把持物体/道具ごとにグルーピングされ, PBの空間が把持物体/道具のダイナミクスごとに自己組織化していることがわかる.
  学習時には用いなかった, Cylinderと同じ形状であるがGripperと同程度の半径を持つ物体Cylinder-L (\figref{figure:hand-dynamics}の(c)上部)についてデータを収集しPBのみ更新すると, PBの空間でCylinderとGripperの中間あたりにCylinder-LのPBが配置され, 学習していない把持物体/道具についてもPBの空間が汎化していることがわかる.
  また, \figref{figure:hand-dynamics}の点線で表したPBの軌跡から, 把持物体/道具を変化させながらランダム動作を行うと同時にPBのオンライン学習を実行すると, 現在のPB $\bm{p}$が徐々に現在の把持物体/道具に関する学習されたPB $\bm{p}_{k}$に近づいていく.
  つまり, そのダイナミクスから把持物体認識を行うことが可能であることがわかる.
  この際の平均ステップ数は500であり, 約100秒で適応することがわかる.

  \figref{figure:hand-dynamics}の(d)にはDPMPBを使った把持安定化制御実験を示す.
  \secref{subsec:control}における$h_{loss}$を以下のように設定することで, 物体把持初期の把持状態を外力に抗って保つことが可能である.
  \begin{align}
    h_{loss}(\bm{s}^{pred}_{seq}) = &||\bm{w}_{1}\otimes(\bm{F}^{pred}_{seq}-\bm{F}^{init}_{seq})||_{2} \nonumber\\
    &+w_{2}||\bm{\theta}^{pred}_{seq}-\bm{\theta}^{init}_{seq}||_{2} + w_{3}||\bm{l}^{pred}_{seq}-\bm{l}^{init}_{seq}||_{2} \label{eq:hand-control-loss}
  \end{align}
  なお, $\{\bm{F}, \bm{\theta}, \bm{l}\}^{init}_{seq}$は物体把持初期の$\{\bm{F}, \bm{\theta}, \bm{l}\}$, $\{\bm{w}_{1}, w_{2}, w_{3}\}$は重みの係数, $||\cdot||_{2}$はL2ノルム, $\otimes$はアダマール積を表す.
  $\bm{w}_{1}$は, 接触センサや筋張力センサの値がマイナス方向には0までしか変化しないが, プラス方向には定格まで大きく変化するという異方性を考慮し, $\bm{F}^{pred}_{seq}[i] \geq \bm{F}^{ref}_{seq}[i])$のとき$\bm{w}_{1}[i]=1.0$, それ以外のとき$\bm{w}_{1}[i]=w_4 (w_4 > 1.0)$という形に設定している.
  グラフから, ハンマーで机を叩く際に, この把持安定化制御を入れない場合は$h_{loss}$で表現された$L$が徐々に大きくなり, 入れた場合は大きな外力に対してそれに抗い$L$を小さく保つことが可能であった.

  \figref{figure:hand-dynamics}の(e)にはDPMPBを使った異常検知の応用実験を示す.
  Noneにおいて学習された$\bm{p}_{k}$を現在の$\bm{p}$として設定し, 物体/道具を把持することでダイナミクスが変化し, ネットワークに予測誤差が生じる.
  これを\secref{subsec:anomaly-detection}の異常検知により捉えることで, 把持検知を行うことが可能になる.
  なお, PBの空間でよりNoneから遠い物体ほど, 把持した際に異常度$d$が大きく変化しやすい.
}%

\begin{figure}[t]
  \centering
  \includegraphics[width=0.99\columnwidth]{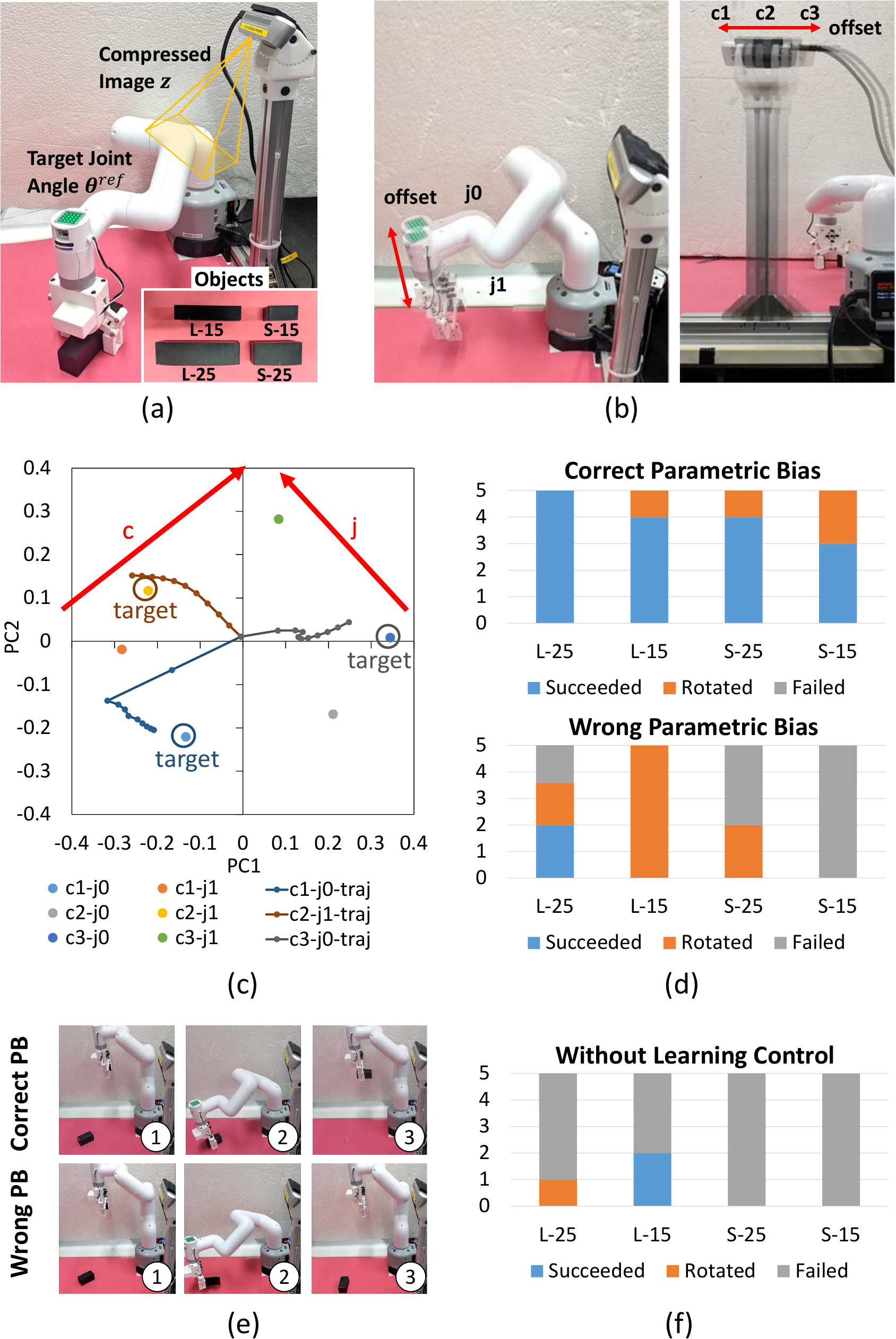}
  \vspace{-3.0ex}
  \caption{Experiment of visual feedback of low-rigidity robot considering temporal body change \cite{kawaharazuka2022vservoing}. (a) shows the sensors and actuators of the low-rigidity robot MyCobot, (b) shows the changed offset values of joint angle and camera arrangement to express temporal change in robot configuration, (c) shows the trained parametric bias and its trajectory when conducting online update of PB, (d) shows the result of visual feedback experiment when using correct or wrong parametric bias, (e) shows examples of the visual feedback motion of (d), and (f) shows the result of general grasping control using point cloud and inverse kinematics.}
  \label{figure:visual-feedback}
  \vspace{-3.0ex}
\end{figure}

\subsection{Visual Feedback of Low-Rigidity Robot Considering Temporal Body Change} \label{subsec:visual-feedback}
\switchlanguage%
{%
  In this section, we deal with a visual feedback model of low-rigidity plastic-made axis-driven robot MyCobot as a modeling difficulty.
  As a temporal model change, we deal with the physical changes in robot state such as aging and configuration changes of the robot.
  For sensor state and control input, we set $\bm{s}=\bm{z}$ and $\bm{u}=\bm{\theta}^{ref}$.
  Note that, as shown in (a) of \figref{figure:visual-feedback}, $\bm{z}$ is the image compressed by AutoEncoder, and $\bm{\theta}^{ref}$ is the target joint angle $\in\mathbb{R}^{7}$ (the grasping state is represented by a binary value of 0 or 1).
  We use four objects \{L-25, L-15, S-25, S-15\} to be grasped, which are combinations of the length of the object (L or S) and the width of the object (15 mm or 25 mm).
  In (b) of \figref{figure:visual-feedback}, the aging of the joints is represented by \{j0, j1\} and the configuration change of the camera position is represented by \{c1, c2, c3\}.
  Since it is difficult to quantitatively evaluate the changes in the robot state over time, we artificially create and experimentally test these changes by assuming that j1 is the case where all joints are offset by 2 deg with respect to j0, and that \{c1, c2, c3\} is the case where the camera position is shifted by 10 mm.
  This experiment is difficult in that the robot with a flexible body needs to perform imitation behaviors while responding to time-varying body states.
  We collected data by a series of motions (approaching, grasping an object, and returning) while changing the robot state.
  In this process, since a robot with low rigidity cannot grasp an object as intended if it simply reaches for the recognized object, the robot itself first randomly places the object and then reaches out to the same position to grasp it.
  By repeating this procedure, data is autonomously collected for visual feedback without human instruction.
  DPMPB with 2-dimensional PB was trained using 120 datasets with about 17 steps each (about 2040 steps in total)
  Here, $L_{1}=0.088$ and $L_{2}=0.022$, and since both values are low and $\bm{u}$ can be inferred, the network structure is CTM.
  The final loss when training was 0.013.

  (c) of \figref{figure:visual-feedback} shows the placement of the trained PBs $\bm{p}_{k}$.
  We can see that the space of PB is self-organized along the axes of \{c1, c2, c3\} and \{j0, j1\}.
  In addition, from the trajectory of PB represented as ``-traj'' in (c), when the online learning of PB is performed with the robot state set to \{c1-j0, c2-j1, c3-j0\}, the current robot state can be accurately recognized from its dynamics.
  Here, the average number of data steps used for online update is 17, which indicates that the adaptation takes about 17 seconds.

  In (d) and (e) of \figref{figure:visual-feedback}, we show the results of visual feedback experiments using DPMPB.
  Four objects are placed at different positions and the robot grasps them five times each by using visual feedback.
  Here, "Succeeded" means that the robot succeeded in grasping the object, "Rotated" means that the robot succeeded in grasping the object but the parallel gripper hit the edge of the object and the object rotated, and "Failed" means that the robot failed to grasp the object.
  The robot state is c2-j0, and we compare the case where the trained PB for the same state is used (Correct) and the case where the trained PB for the wrong state c3-j1 is used (Wrong).
  As shown in (e), we can see that the visual feedback motion changes depending on the value of PB.
  From (d), we can see that in the case of Correct, all grasps Succeeded or Rotated, while in the case of Wrong, the probability of failure increases, indicating that the recognition of the current robot state is important.
  In both cases, the larger object is more likely to be grasped successfully even if its position is slightly misaligned.

  In (f) of \figref{figure:visual-feedback}, we show the results of a general object grasping experiment without any learning control.
  A point cloud is obtained from a depth camera, the object is extracted from plane detection and Euclidean clustering, and the object is grasped by solving inverse kinematics.
  The robot body is approximated as a rigid body and the camera position on CAD (Computer-Aided Design) is used.
  The success rate of grasping is very low for all objects, and most of the movements do not even touch the object.
}%
{%
  本研究はモデル化困難性として, 低剛性樹脂製の軸駆動型ロボットであるMyCobotの視覚フィードバックモデルを扱う.
  また, 逐次的モデル変化として, ロボットの経年劣化やコンフィギュレーション変化等の状態変化を扱う.
  感覚と運動については, $\bm{s}=\bm{z}$, $\bm{u}=\bm{\theta}^{ref}$とした.
  なお, \figref{figure:visual-feedback}の(a)に示すように, $\bm{z}$はAutoEncoderにより圧縮された画像, $\bm{\theta}^{ref}$は指令関節角度$\in\mathbb{R}^{7}$ (把持状態については0または1のバイナリで表現する)である.
  扱う物体については, 物体の長さ(LまたはS)と物体の幅(15 mmまたは25 mm)の組み合わせである4つの物体\{L-25, L-15, S-25, S-15\}を用いる.
  \figref{figure:visual-feedback}の(b)では, 簡易的に関節の経年劣化を\{j0, j1\}, カメラ位置のコンフィギュレーション変化を\{c1, c2, c3\}で表現している.
  時間経過におけるロボットの状態変化を定量的に評価するのは難しいため, j1はj0に対して全ての関節に2 degのオフセットを乗せた場合, \{c1, c2, c3\}はカメラ位置を10 mmずつずらした場合として人為的にそれらの変化状態を作成し実験を行う.
  本実験は, 柔軟身体を持つロボットが, 時間的に変化する身体状態に対応しながら模倣動作を行う必要がある点で難しい.
  ロボット状態を変化させながら, 物体に近づき把持して戻る一連の動作によりデータを収集する.
  この際, 低剛性なロボットは単に認識した物体に手を伸ばしても意図したようその物体を把持できないため, 把持している物体をロボットがランダムに置き, 同じ位置に手を伸ばして把持する動作を繰り返すことで, 人間の教示なしに自律的に視覚フィードバックに用いるデータを収集する.
  PBを2次元として, 17ステップのデータを120回, 計2040ステップのデータを用いて学習している.
  この際, $L_{1}=0.088$, $L_{2}=0.022$となり, どちらの値も低く$\bm{u}$も推論できているため, ネットワーク構造はCTMとなった.
  なお, 最終的な訓練時の推論誤差は0.013であった.


  \figref{figure:visual-feedback}の(c)には学習されたPB $\bm{p}_{k}$の配置を示す.
  PBが\{c1, c2, c3\}と\{j0, j1\}の軸に沿って規則的に自己組織化していることがわかる.
  また, \figref{figure:visual-feedback}に``-traj''として表したPBの軌跡から, ロボット状態を\{c1-j0, c2-j1, c3-j0\}に設定して動作を行った際にPBのオンライン学習を実行すると, そのダイナミクスから現在のロボット状態が正確に認識されることがわかった.
  この際の平均ステップ数は17であり, 約17秒で適応することがわかる.

  \figref{figure:visual-feedback}の(d)と(e)にはDPMPBを使った視覚フィードバック実験の結果を示す.
  4つの物体をそれぞれ別の位置に配置し, これを視覚フィードバックによりそれぞれ5回ずつ把持する実験を行う.
  ここで, Succeededは把持に成功した状態, Rotatedは把持には成功しているがグリッパが物体の端に当たり物体が回転してしまった状態, Failedは物体把持に失敗した状態を表す.
  また, ロボット状態はc2-j0とし, 同様の状態について学習されたPBを使った場合(Correct)と, c3-j1の状態について学習された間違ったPBを使った場合(Wrong)を比較している.
  (e)に示すように, PBの値によってその動作が変化していることがわかる.
  (d)から, Correctの場合はSucceededまたはRotatedで全て把持に成功しているのに対して, Wrongの場合は, 把持に失敗する確率が高まり, 現在のロボット状態の認識が重要であることがわかる.
  また, どちらの場合も, より大きな物体の方が多少位置がズレても把持に成功する.

  最後に, \figref{figure:visual-feedback}の(f)に, 一切の学習制御を用いない一般的な物体把持実験の結果を示す.
  深度カメラからポイントクラウドを取得し, 平面検出とEuclidean Clusteringから物体を切り出し, 逆運動学を解くことで物体を把持する.
  この際は剛体近似されたロボットとCAD上のカメラ位置等を用いている.
  どの物体についても把持の成功率は非常に低く, ほとんどの動作で物体に触ることすら出来ていない.
}%

\begin{figure}[t]
  \centering
  \includegraphics[width=0.99\columnwidth]{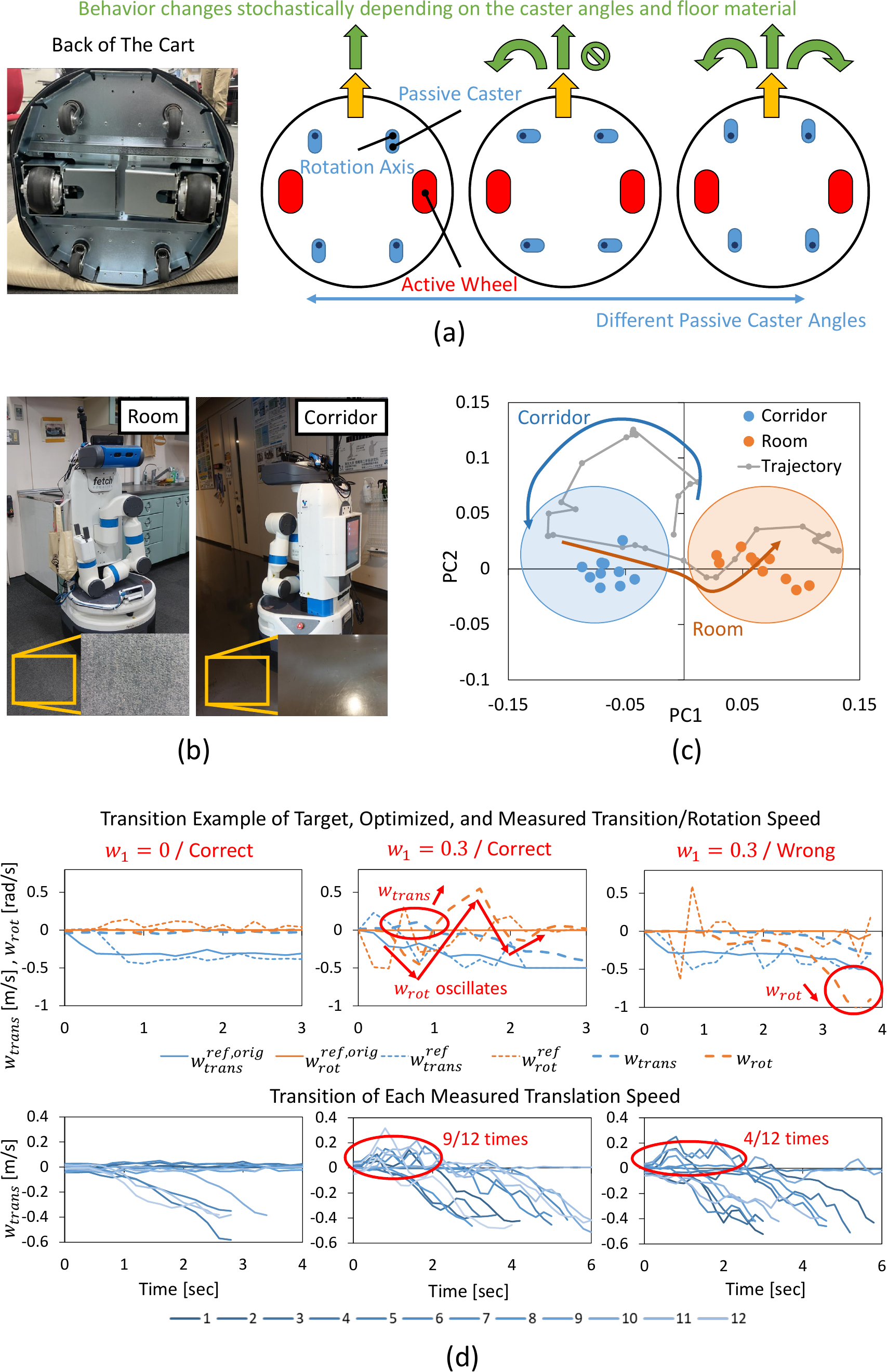}
  \vspace{-3.0ex}
  \caption{Experiment of stable control of wheeled robot considering floor change \cite{kawaharazuka2021fetch}. (a) shows the problem of probabilistic movement of the wheeled robot Fetch, (b) shows the changed floor environment of Room and Corridor, (c) shows the trained parametric bias and its trajectory when conducting online update of PB, (d) shows the result of stable control with variance minimization when setting $w_{1}=0$ / Correct, $w_{1}=0.3$ / Correct, and $w_{1}=0.3$ / Wrong.}
  \label{figure:stable-fetch}
  \vspace{-3.0ex}
\end{figure}

\subsection{Stable Control of Wheeled Robot Considering Floor Change} \label{subsec:stable-fetch}
\switchlanguage%
{%
  In this section, we deal with the sensory-motor model of a wheeled robot Fetch as a modeling difficulty.
  As a temporal model change, we deal with the change in the operating environment of the wheeled robot, i.e., the floor.
  For sensor state and control input, we set $\bm{s}=\bm{w}$ and $\bm{u}=\bm{w}^{ref}$.
  Note that $\bm{w}$ is the current robot velocity obtained by Visual Odometry $\in\mathbb{R}^{2}$ (the translational direction is $w_{trans}$ and the rotational direction is $w_{rot}$), and $\bm{w}^{ref}$ is the target robot velocity.
  As shown in (a) of \figref{figure:stable-fetch}, the wheeled base of Fetch has two active wheels and four passive casters.
  When moving backwards, the motion of the base may change stochastically depending on the direction of the casters, which is characteristic of the old Fetch model (it is possible to move forward accurately due to the suspension).
  In order to deal with the stochastic state transitions in this experiment, the network output in STM is changed to the mean $\bm{s}$ and variance $\bm{v}$ of the states, and is trained by a loss function based on the following maximum likelihood estimation,
  \begin{align}
    P(s^{k}_{i, t}|D_{k, 1:t-1}, W, \bm{p}_{k}) &= \frac{1}{\sqrt{2\pi \hat{v}_{i, t}}}\exp{\left(-\frac{(\hat{s}^{k}_{i, t}-s^{k}_{i, t})^{2}}{2\hat{v}_{i, t}}\right)} \nonumber\\
    L_{likelihood}(W, \bm{p}_{1:K}|D_{train}) &= \prod^{K}_{k=1}\prod^{T_{k}}_{t=1}\prod^{N_{s}}_{i=1}P(s^{k}_{i, t}|D_{k, 1:t-1}, W, \bm{p}_{k}) \nonumber\\
    L_{train} &= -\log(L_{likelihood}) \label{eq:fetch-loss}
  \end{align}
  where $P$ is the probability density function, $\{s, v\}_{i}$ is $\{s, v\}$ of the $i$-th sensor, $D_{k, 1:t-1}$ is the data of $D_k$ in $[1, t-1]$, $\{\hat{s}, \hat{v}\}$ is the predicted value by using $D_{k, 1:t-1}$, $\bm{W}$, and $\bm{p}_{k}$, $\bm{p}_{1:K}$ denotes a vector summarizing $\bm{p}_{k}$ in $1 \leq k \leq K$, and $N_s$ is the dimension of $\bm{s}$.
  $L_{likelihood}$ represents the likelihood function for $\bm{W}$ and $\bm{p}$ given $D_{train}$, and we consider maximizing it and transform it to the problem of minimizing the loss $L_{train}$ by taking -log.
  In (b) of \figref{figure:stable-fetch}, we prepare two kinds of environments, Room with high friction and Corridor with low friction.
  This experiment is difficult in that it is necessary to learn a stochastic and difficult-to-modelize behavior depending on the operating environment, and to control the body while recognizing the environment and considering the variance of the motion.
  We collected data by random control input while changing the floor.
  DPMPB with 2-dimensional PB was trained using 9 datasets with about 600 steps each (about 5400 steps in total).
  Here, $L_{1}=0.112$ and $L_{2}=0.949$, and since $\bm{y}_{2}=\bm{u}$ is removed from the output by setting $L_{thre}=0.3$, the network structure is STM.
  The final loss when training was -0.040 (which can be negative because of the use of \equref{eq:fetch-loss}).

  In (c) of \figref{figure:stable-fetch}, we show the arrangement of the trained PBs $\bm{p}_{k}$.
  We can see that PBs are neatly divided into Room and Corridor, and that the space of PB is self-organized.
  From the trajectory of PB represented as ``Trajectory'' in (c), by updating PB online, we can see that when the robot is in Corridor, the current $\bm{p}$ approaches $\bm{p}_{k}$ trained in Corridor, and when the robot is in Room, the current $\bm{p}$ approaches $\bm{p}_{k}$ trained in Room, indicating that the operating environment can be correctly recognized from its dynamics.
  Here, the average number of data steps used for online update is 150, which indicates that the adaptation takes about 30 seconds.

  (d) of \figref{figure:stable-fetch} shows a stabilization control experiment based on variance minimization using DPMPB.
  The control can be achieved by setting $h_{loss}$ in \secref{subsec:control} as follows,
  \begin{align}
    h_{loss}(\bm{s}^{pred}_{seq}, \bm{v}^{pred}_{seq}) = ||\bm{s}^{ref}_{seq}-\bm{s}^{pred}_{seq}||_{2} + w_{1}||\bm{v}^{pred}_{seq}||_{2}\label{eq:fetch-control-loss}
  \end{align}
  where $w_{1}$ denotes the constant weight.
  This corresponds to the control that makes the current state closer to the target state while minimizing the variance of the state.
  In this experiment, $\bm{s}^{ref}$ is represented as $\bm{w}^{ref, orig}$, and the optimized value $\bm{u}^{opt}$ is represented as $\bm{w}^{ref}$.
  The effectiveness of this control is verified by moving the robot backward after one rotation.
  The passive caster faces perpendicularly to the active wheel after one rotation, and when the robot moves backward in this state, the motion often gets stuck.
  This experiment is conducted in Room, and three cases are compared: $w_{1}=\{0, 0.3\}$ for the case of using $\bm{p}$ updated in Room (Correct), and $w_{1}=0.3$ for the case of using $\bm{p}$ updated in Corridor (Wrong).
  For each condition, the transitions of $\bm{w}^{ref, orig}$, $\bm{w}^{ref}$, and $\bm{w}$ are shown in the upper figure of (d) of \figref{figure:stable-fetch}.
  When $w_{1}=0$, there is no significant difference between $\bm{w}^{ref, orig}$ and $\bm{w}^{ref}$, but $w_{trans}$ remains 0 and the motion is completely stucked.
  In contrast, in the case of $w_{1}=0.3$ / Correct, $\bm{w}^{ref, orig}$ and $\bm{w}^{ref}$ differ significantly, and $w_{trans}$ starts to move after about 1.5 seconds.
  A characteristic of this condition is that $w_{trans}$ first moves forward to change the direction of the caster, and $\bm{w}_{rot}$ is moved while oscillating to prevent it from getting stuck.
  In the case of $w_{1}=0.3$ / Wrong, $w_{trans}$ also moves.
  In this condition, compared to $w_{1}=0.3$ / Correct, $w_{trans}$ is not moved forward, but $w_{trans}$ is directly moved backward.
  In addition, there are many cases where $w_{rot}$ moves significantly in the same direction as that of the previous rotation; that is, it moves backward while rotating.
  $w_{trans}$ for 12 trials is shown in the lower figure of (d) in \figref{figure:stable-fetch}.
  In the case of $w_{1}=0$, the motion is stuck in more than half of the trials, while in the case of $w_{1}=0.3$, the motion in the backward direction succeeds in all but one or two trials.
  In the case of $w_{1}=0.3$, the number of trials in which the robot moves backward after moving forward once is 9 out of 12 in the case of Correct, and 4 out of 12 in the case of Wrong.
}%
{%
  本研究はモデル化困難性として, 台車型ロボットFetchの感覚-運動モデルを扱う.
  また, 逐次的モデル変化として, 台車型ロボットの動作環境, すなわち床の変化を扱う.
  感覚と運動については, $\bm{s}=\bm{w}$, $\bm{u}=\bm{w}^{ref}$とした.
  なお, $\bm{w}$はVisual Odometryにより得られた現在の車体速度$\in\mathbb{R}^{2}$ (並進方向を$w_{trans}$, 回転方向を$w_{rot}$とする), $\bm{w}^{ref}$は指令車体速度である.
  \figref{figure:stable-fetch}の(a)に示すように, Fetchの台車には2つの能動輪と4つの受動キャスタがあり, このキャスタの向き次第で上手く後ろに進めずに動作が確率的に変化してしまうという問題に対処する(なお, 前進はサスペンションの関係か容易に可能である. また, この問題は古いFetchの型に特徴的である).
  本実験では確率的状態遷移を扱うため, STMにおけるネットワーク出力を状態の平均$\bm{s}$と分散$\bm{v}$として, 以下の最尤推定に基づく損失関数により学習を行う.
  \begin{align}
    P(s^{k}_{i, t}|D_{k, 1:t-1}, W, \bm{p}_{k}) &= \frac{1}{\sqrt{2\pi \hat{v}_{i, t}}}\exp{\left(-\frac{(\hat{s}^{k}_{i, t}-s^{k}_{i, t})^{2}}{2\hat{v}_{i, t}}\right)} \nonumber\\
    L_{likelihood}(W, \bm{p}_{1:K}|D_{train}) &= \prod^{K}_{k=1}\prod^{T_{k}}_{t=1}\prod^{N_{s}}_{i=1}P(s^{k}_{i, t}|D_{k, 1:t-1}, W, \bm{p}_{k}) \nonumber\\
    L_{train} &= -\log(L_{likelihood}) \label{eq:fetch-loss}
  \end{align}
  ここで, $P$は確率密度関数, $\{s, v\}_{i}$は$i$番目のセンサの$\{s, v\}$, $D_{k, 1:t-1}$は$[1, t-1]$の間の$D_k$のデータ, $\{\hat{s}, \hat{v}\}$はデータ$D_{k, 1:t-1}$, 重み$\bm{W}$, $\bm{p}_{k}$を使って予測された$\{s, v\}$の値, $\bm{p}_{1:K}$は$1 \leq k \leq K$の$\bm{p}_{k}$をまとめたベクトルを表す.
  $L_{likelihood}$は $D_{train}$が与えられたときの$\bm{W}$, $\bm{p}$に関する尤度関数を表し, これを最大化する問題を考え, -logをとることで損失$L_{train}$の最小化問題に落とし込む.
  \figref{figure:stable-fetch}の(b)では, 2種類の動作環境として, 摩擦の強いRoom, 摩擦の少ないCorridorを用意した.
  本実験は, 動作環境に依存した確率的でモデル化の難しい挙動を学習し, 動作環境を認識したうえで動きの分散を考慮しつつ身体を制御する必要がある点で難しい.
  床状態を変化させながら, 車体速度のランダム入力によりデータを収集する.
  PBを2次元として, 約600ステップのデータを9回, 計約5400ステップのデータを用いて学習している.
  この際, $L_{1}=0.112$, $L_{2}=0.949$となり, $L_{thre}=0.3$と設定することで, $\bm{y}_{2}=\bm{u}$が出力から削除されSTMとなった.
  なお, 最終的な訓練時の推論誤差は-0.40であった(\equref{eq:fetch-loss}を用いているため負になり得る).


  \figref{figure:stable-fetch}の(c)には学習されたPB $\bm{p}_{k}$の配置を示す.
  PBがRoomとCorridorで綺麗に分かれ, PBの空間が自己組織化していることがわかる.
  また, \figref{figure:stable-fetch}に``Trajectory''として表したPBの軌跡から, PBのオンライン学習を実行すると, ロボットがCorridorにいるときは現在の$\bm{p}$がCorridorにおいて学習された$\bm{p}_{k}$周辺へ, RoomにいるときはRoomにおいて学習された$\bm{p}_{k}$周辺に動き, 動作環境がそのダイナミクスから正しく認識可能なことがわかる.
  この際の平均ステップ数は150であり, 約30秒で適応することがわかる.

  \figref{figure:stable-fetch}の(d)はDPMPBを使った分散最小化に基づく安定化制御実験を示している.
  \secref{subsec:control}における$h_{loss}$を以下のように設定することで, 安定した制御を行うことができる.
  \begin{align}
    h_{loss}(\bm{s}^{pred}_{seq}, \bm{v}^{pred}_{seq}) = ||\bm{s}^{ref}_{seq}-\bm{s}^{pred}_{seq}||_{2} + w_{1}||\bm{v}^{pred}_{seq}||_{2}\label{eq:fetch-control-loss}
  \end{align}
  なお, $w_{1}$は重みの係数を表す.
  これは, 動作の分散を最小化しつつ現在車体速度を指令値に近づける制御に相当する.
  本実験では, $s^{ref}$を$w^{ref, orig}$, 最適化されて最終的に実機送る値$\bm{u}^{opt}$を$w^{ref}$と表現する.
  本制御の有効性を, ロボットを一回転させた後に後ろに進む動作により検証する.
  一回転させると受動キャスターが能動輪に対して垂直に向き, その状態で後ろに進むと, 動作がスタックしてしまうことが多い.
  本実験はRoomで行い, Roomにおいて更新した$\bm{p}$を使う場合(Correct)について$w_{1}=\{0, 0.3\}$を, Corridorにおいて更新した$\bm{p}$を使う場合(Wrong)について$w_{1}=0.3$の場合の三種類について比較する.
  それぞれの条件における, 代表的な動作の一例について$\bm{w}^{ref, orig}$, $\bm{w}^{ref}$, $\bm{w}$の遷移を\figref{figure:stable-fetch}の上図に示す.
  $w_{1}=0$の場合には, $\bm{w}^{ref, orig}$と$\bm{w}^{ref}$の間に大きな違いはないが, $w_{trans}$は0のままで完全に動作がスタックしている.
  それに対して, $w_{1}=0.3$ / Correctの場合は$\bm{w}^{ref, orig}$と$\bm{w}^{ref}$が大きく異なり, $w_{trans}$も1.5秒後程度から動き始めている.
  この条件に特徴的なのは, $w_{trans}$が最初前方に動くことでキャスターの向きを変化させること, そして$\bm{w}_{rot}$を振動させることでスタックしないように動かしている点である.
  $w_{1}=0.3$ / Wrongの場合も同様に$w_{trans}$を動作させることができている.
  この条件では, Correctに比べ, $w_{trans}$を前方に動かさず, 直接後方に対して指令値を送る動作が多い.
  また, $w_{rot}$が前の回転時の動作と同じ方向に大きく動く, つまり回転しながら後方に下がる場合が多い.
  12回の試行について$w_{trans}$を\figref{figure:stable-fetch}の(d)の下図に示す.
  $w_{1}=0$の場合は半分以上の試行で動作がスタックしているのに対して, $w_{1}=0.3$の場合は1,2回の試行を除いて後進方向への動作に成功している.
  なお, $w_{1}=0.3$で一度前進してから後進するケースはCorrectの場合は9/12回, Wrongの場合は4/12回であった.
}%

\begin{figure}[t]
  \centering
  \includegraphics[width=0.99\columnwidth]{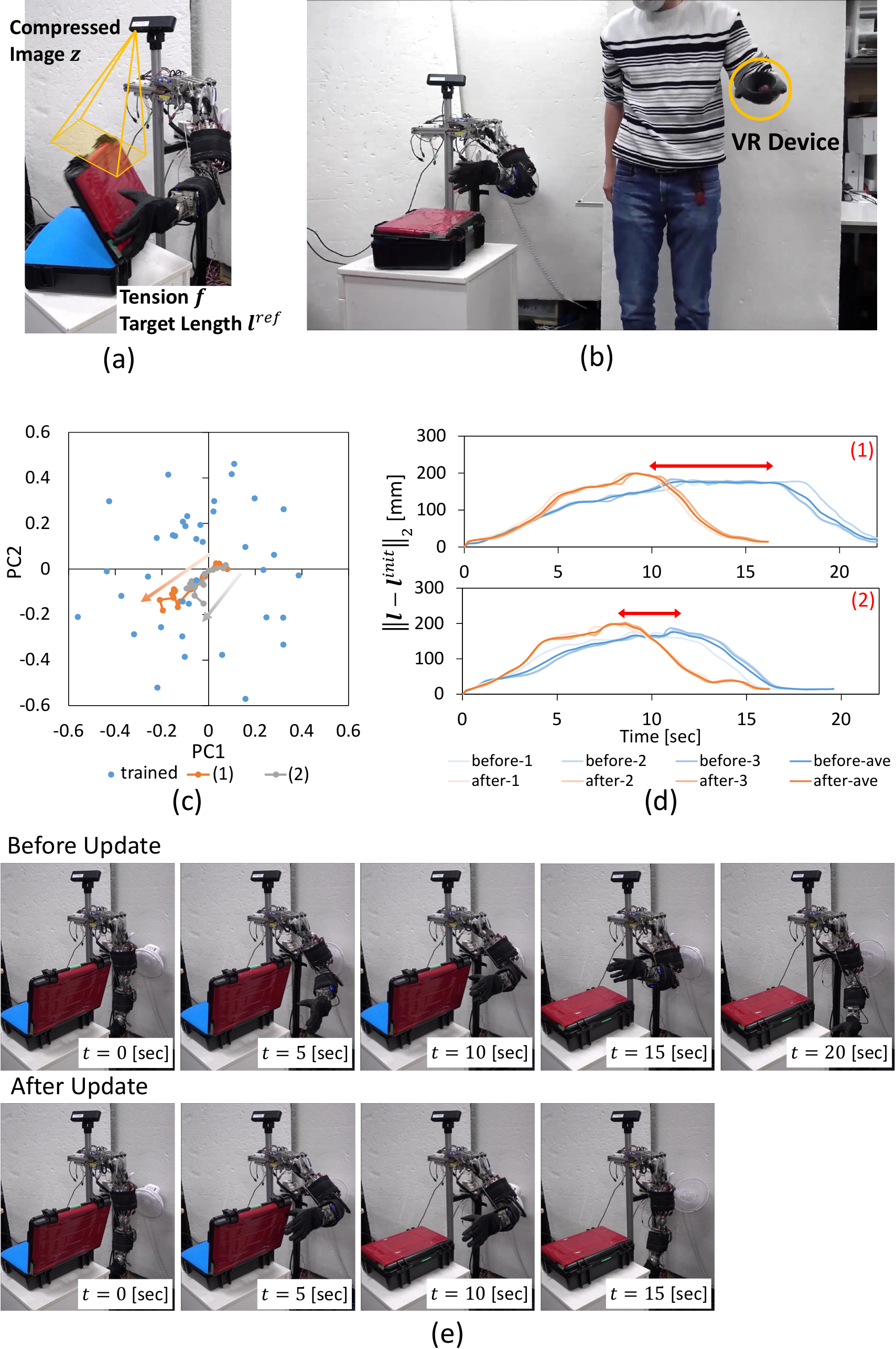}
  \vspace{-3.0ex}
  \caption{Experiment of imitation learning considering motion style \cite{kawaharazuka2021imitation}. (a) shows the sensors and actuators of the musculoskeletal humanoid MusashiLarm, (b) shows the data collection setting using VR device, (c) shows the trained parametric bias and its trajectory when conducting online update of PB, (d) shows the trajectory of $||\bm{l}-\bm{l}^{init}||_{2}$ when maximizing or not maximizing the muscle length velocity, and (e) shows the imitated motion of closing a box before and after the update of PB.}
  \label{figure:imitation-learning}
  \vspace{-3.0ex}
\end{figure}

\subsection{Imitation Learning Considering Motion Style} \label{subsec:imitation-learning}
\switchlanguage%
{%
  In this section, we deal with a human imitation model of a musculoskeletal humanoid MusashiLarm with a flexible body as a modeling difficulty.
  As a temporal model change, we deal with the variability of human motion, i.e., change in motion style.
  This represents, for example, motion variations as different elbow positions or different movement speeds when performing a certain task.
  For sensor state and control input, we set $\bm{s}=\{\bm{z}, \bm{f}\}$ and $\bm{u}=\bm{l}^{ref}$.
  Note that, as shown in (a) of \figref{figure:imitation-learning}, $\bm{z}$ is the image compressed by AutoEncoder, and $\{\bm{f}, \bm{l}^{ref}\}$ is \{muscle tension, target muscle length\} of the muscles related to the arm $\in\mathbb{R}^{10}$ (5 DOFs of the shoulder and elbow are used).
  As shown in (b) of \figref{figure:imitation-learning}, a human moves MusashiLarm to close a box on the desk using a VR device, and the difference in the motion style is embedded in PB.
  This experiment is difficult in that it is necessary for a robot with a flexible body to learn to imitate human behaviors while changing its motion style.
  We collected data by performing a box-closing motion while changing the position and angle of the box.
  DPMPB with 3-dimensional PB was trained using 30 datasets with about 100 steps each (about 3000 steps in total)
  Here, $L_{1}=0.067$, $L_{2}=0.050$, and $L_{3}=0.013$, and since all values are low and $\bm{u}$ can be inferred, the network structure is CTM.
  The final loss when training was 0.017.

  It is possible to realize the desired motion style by actively changing the motion style embedded in PB.
  In this study, we update $\bm{p}$ online to maximize the motion speed using the obtained data $D$, by setting the loss function as follows,
  \begin{align}
    L_{update} = ||\bm{s}^{data}_{2:T}-\bm{s}^{pred}_{2:T}||_2 + w_{1}||\bm{u}^{pred}_{3:T}-\bm{u}^{pred}_{2:T-1}||_{2} \label{eq:style-loss}
  \end{align}
  where $\{\bm{s}, \bm{u}\}_{1:T}$ is $\{\bm{s}, \bm{u}\}$ in $[1, T]$ ($T$ represents the number of time steps in $D$), $\{\bm{s}, \bm{u}\}^{data}$ is the data contained in $D$, and $w_{1}$ is the constant weight (in this study, $w_{1}<0$ for the speed maximization).
  $\{\bm{s}, \bm{u}\}^{data}_{1}$ is fed into the network and $\{\bm{s}, \bm{u}\}^{pred}$ is inferred in an autoregressive manner.
  We maximize the velocity of $\bm{u}$ while imposing a slight constraint on $\bm{s}$.

  In (c) of \figref{figure:imitation-learning}, we show the arrangement of the trained PBs $\bm{p}_{k}$ and the trajectories (1) and (2) of $\bm{p}$ during online learning based on this speed maximization.
  Although (1) and (2) show results for different box angles, $\bm{p}$ transitions in the same direction in both cases, indicating that the space of PB is self-organized to represent the motion style.
  Here, the average number of data steps used for online update is 100, which indicates that the adaptation takes about 20 seconds.
  In (d) of \figref{figure:imitation-learning}, the transition of the change in $\bm{l}$ from the initial muscle length $\bm{l}^{init}$ is shown.
  before-\{1, 2, 3\} denotes the three trials before the update of $\bm{p}$, after-\{1, 2, 3\} denotes the three trials after the update of $\bm{p}$, and \{before, after\}-ave denotes their average.
  It can be seen that the behavior after the update of $\bm{p}$ converges faster than before, and at the same time, the behavior is reproducible.
  (e) of \figref{figure:imitation-learning} shows snapshots of the motion in (1), where the robot completes the box-closing operation more than 5 seconds faster after updating $\bm{p}$.
}%
{%
  本研究はモデル化困難性として, 柔軟身体を持つ筋骨格ヒューマノイドMusashiLarm \cite{kawaharazuka2019musashi}の模倣動作モデルを扱う.
  また, 逐次的モデル変化として, 本節は特殊な系として, 人間の動作のばらつき, つまり動作スタイル変化を扱う.
  これは例えば, あるタスクを行う際に肘の位置が異なったり, 動作のスピードが異なったりといった動きのばらつきを表している.
  感覚と運動については, $\bm{s}=\{\bm{z}, \bm{f}\}$, $\bm{u}=\bm{l}^{ref}$とした.
  なお, \figref{figure:imitation-learning}の(a)に示すように, $\bm{z}$はAutoEncoderにより圧縮された画像, $\{\bm{f}, \bm{l}^{ref}\}$は腕(肩と肘の5自由度)に関係する筋の\{筋張力, 指令筋長\}$\in\mathbb{R}^{10}$である.
  \figref{figure:imitation-learning}の(b)のように人間がVRデバイスを使ってMusashiLarmにより箱を閉じる動作を行い, その際の動作スタイルの違いがPBに埋め込まれる.
  本実験は, 柔軟身体を持つロボットがその動作スタイルを変化させながら模倣学習を行う点で難しい.
  箱の位置や角度を変化させながら箱閉め動作を行いデータを収集する.
  PBを3次元として, 約100ステップのデータを30回, 計約3000ステップのデータを用いて学習している.
  この際, $L_{1}=0.067$, $L_{2}=0.050$, $L_{3}=0.013$となり, どの値も低く$\bm{u}$も推論できているため, ネットワーク構造はCTMとなった.
  なお, 最終的な訓練時の推論誤差は0.017であった.

  PBに埋め込んだ動作スタイルを能動的に変化させ, 所望の動作スタイルを実現することが可能である.
  本研究では, 以下のように損失関数を設定して, 得られたデータ$D$をもとに$\bm{p}$のみオンラインで更新することで, 動作における速度最大化を行った.
  \begin{align}
    L_{update} = ||\bm{s}^{data}_{2:T}-\bm{s}^{pred}_{2:T}||_2 + w_{1}||\bm{u}^{pred}_{3:T}-\bm{u}^{pred}_{2:T-1}||_{2} \label{eq:style-loss}
  \end{align}
  なお, $\{\bm{s}, \bm{u}\}_{1:T}$は$[1, T]$の間の$\{\bm{s}, \bm{u}\}$ ($T$はデータ$D$のステップ数を表す), $\{\bm{s}, \bm{u}\}^{data}$は$D$に含まれるデータ, $w_{1}$は重みの係数(今回は速度最大化のため$w_{1}<0$)を表す.
  ネットワークに, $\{\bm{s}, \bm{u}\}^{data}_{1}$を入力して$\{\bm{s}, \bm{u}\}^{pred}$を自己回帰的に推論し, 損失を計算する.
  $\bm{s}$については多少の制約をかけつつ, $\bm{u}$の速度を最大化する.

  \figref{figure:imitation-learning}の(c)には学習されたPB $\bm{p}_{k}$の配置と, この速度最大化に基づくオンライン学習の際の$\bm{p}$の軌跡(1)と(2)を示す.
  (1)と(2)は異なる箱の角度について速度最大化を行っているが, 両者とも$\bm{p}$は同じ方向に遷移し, PBの空間が動作スタイルを表現するよう自己組織化していることがわかる.
  この際の平均ステップ数は100であり, 約20秒の最適化を行った.
  \figref{figure:imitation-learning}の(d)には, 初期筋長$\bm{l}^{init}$からの$\bm{l}$の変化の遷移を示している.
  before-\{1, 2, 3\}は$\bm{p}$の更新前の3回分の試行, after-\{1, 2, 3\}は$\bm{p}$の更新後の3回分の試行, \{before, after\}-aveはそれらの平均を示す.
  $\bm{p}$の更新後は更新前に比べより速く動作が収束していると同時に, 動作には再現性があることがわかる.
  \figref{figure:imitation-learning}の(e)は(1)の動作におけるsnapshotを示しているが, $\bm{p}$の更新により5秒以上速く箱閉め動作を完遂している.
}%

\begin{figure}[t]
  \centering
  \includegraphics[width=0.99\columnwidth]{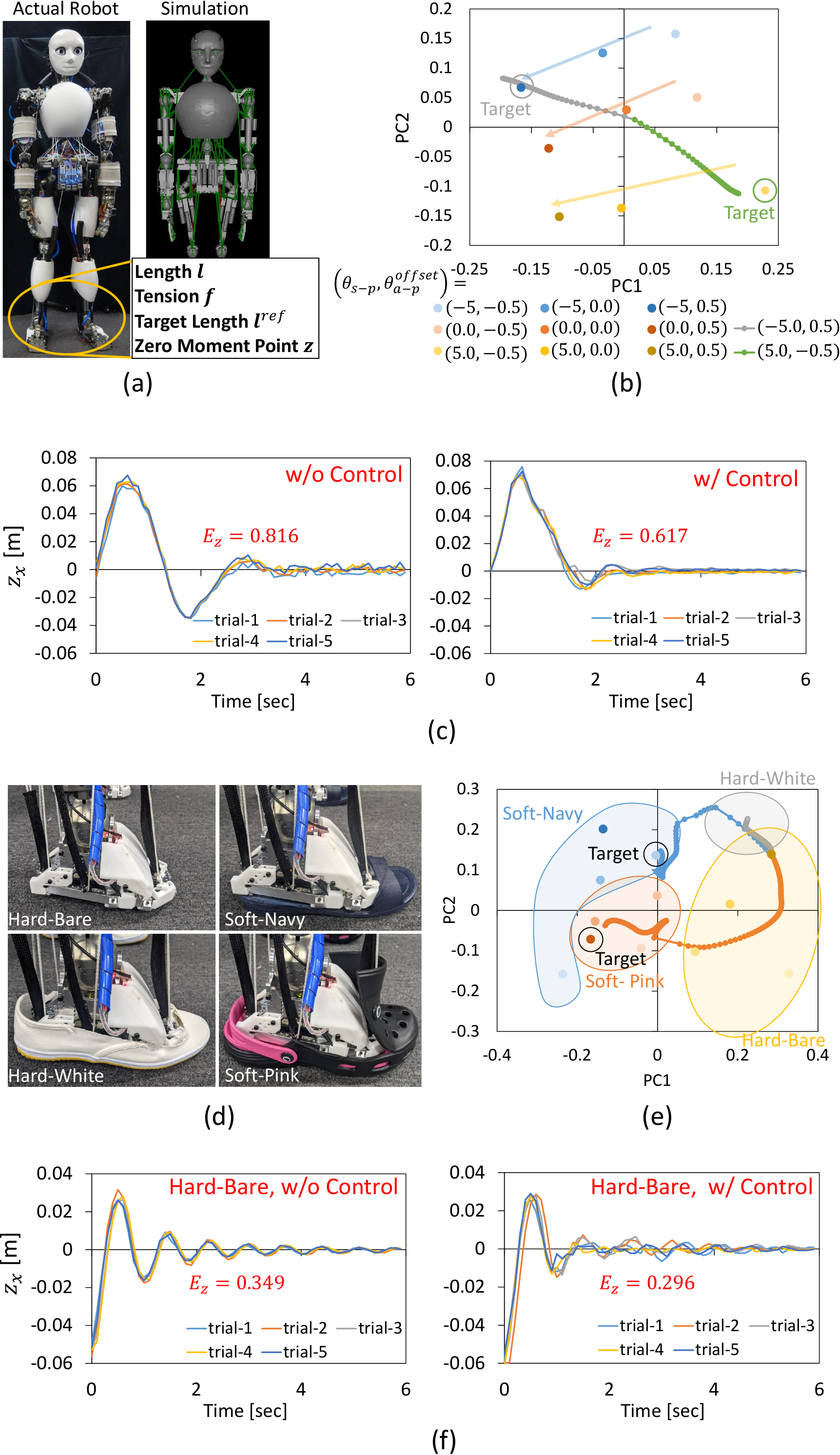}
  \vspace{-3.0ex}
  \caption{Experiment of balance control of a musculoskeletal humanoid considering change in shoes \cite{kawaharazuka2022balance}. (a) shows the sensors and actuators of the musculoskeletal humanoid Musashi, (b) shows the trained parametric bias and its trajectory when conducting online update of PB in simulation, (c) shows the transition of $z_{x}$ when using balance control or not in simulation, (d) shows the shoes worn by the robot, (e) shows the trained parametric bias and its trajectory when conducting online update of PB in the actual robot, (f) shows the transition of $z_{x}$ when using balance control or not in the actual robot.}
  \label{figure:balance-control}
  \vspace{-3.0ex}
\end{figure}

\subsection{Balance Control of a Musculoskeletal Humanoid Considering Change in Shoes} \label{subsec:balance-control}
\switchlanguage%
{%
  In this section, we deal with a balance model of a full-body musculoskeletal humanoid Musashi \cite{kawaharazuka2019musashi} as a modeling difficulty.
  As a temporal model change, we deal with the physical changes in body state that are not included in the balance model, such as changes in shoes and upper body posture.
  For sensor state and control input, we set $\bm{s}=\{\bm{z}, \bm{f}, \bm{l}\}$ and $\bm{u}=\bm{l}^{ref}$.
  Note that, as shown in (a) of \figref{figure:balance-control}, $\bm{z}$ is the zero moment point (zmp) $\in\mathbb{R}^{2}$ ($\{x, y\}$ direction is denoted as $\{z_x, z_y\}$), $\{\bm{f}, \bm{l}, \bm{l}^{ref}\}$ is \{muscle tension, muscle length, target muscle length\} of the muscles related to both ankles $\in\mathbb{R}^{12}$.
  This experiment is difficult in that it is necessary to control the balance of a bipedal humanoid with a flexible body while adapting to changes in its body state.
  Experiments are conducted on the simulation and the actual robot.
  In the simulation, the spine pitch joint $\theta_{s-p}$ and the offset of the ankle pitch joint angle $\theta^{offset}_{a-p}$, which represents calibration deviation, are treated as changes in the body state.
  We collected data while changing the body state to nine combinations of $\theta_{s-p}=\{-5.0, 0.0, 5.0\}$ [deg] and $\theta^{offset}_{a-p}=\{-5.0, 0.0, 5.0\}$ [deg].
  DPMPB with 2-dimensional PB was trained using 9 datasets with about 300 steps each (about 2700 steps in total).
  Here, $L_{1}=0.045$, $L_{2}=0.018$, $L_{3}=0.047$, and $L_{4}=0.296$, and since $\bm{y}_{\{4\}}=\bm{u}$ is removed from the output by setting $L_{thre}=0.2$, the network structure is STM.
  The final loss when training was 0.051.
  In the actual robot, we handle which shoes in (d) of \figref{figure:balance-control} are used and the posture change of the upper body as the changes in body state.
  We collected data in three types of shoes (Hard-Bare, Soft-Pink, and Soft-Navy) and four types of upper body postures (Hard-White is not used for learning).
  Since it is necessary to move the robot within the range where it will not fall down, we gradually increase the random displacement of the control input and periodically change the random width of the input to obtain useful data for balance control.
  DPMPB with 2-dimensional PB was trained using 12 datasets with about 300 steps each (about 3600 steps in total), and the loss when training was 0.074.

  (b) of \figref{figure:balance-control} shows the arrangement of PBs $\bm{p}_{k}$ trained in the simulation, and we can see that the space of PB is neatly self-organized along the axes of $\theta_{s-p}$ and $\theta^{offset}_{a-p}$.
  This is also the case in the actual robot.
  (e) of \figref{figure:balance-control} shows the arrangement of PBs $\bm{p}_{k}$ trained in the actual robot, where $\bm{p}_{k}$ is grouped for each pair of shoes and the space of PB is self-organized.
  When online learning is performed for both (b) and (e), we can see that the current $\bm{p}$ gradually approaches $\bm{p}_{k}$ trained in the current body state, which is represented by ``Target''.
  In addition, the PB of Hard-White, which is not used in the training, is above PBs of Hard-Bare, which may reflect the fact that Hard-White and Hard-Bare, Soft-Navy and Soft-Pink have similar sole hardness.
  Here, the average number of data steps used for online update is 280, which indicates that the adaptation takes about 56 seconds.

  In (c) and (f) of \figref{figure:balance-control}, we show the results of the balance control experiment using DPMPB.
  After updating PB online, we can perform the balance control by setting $h_{loss}$ in \secref{subsec:control} as follows,
  \begin{align}
    h_{loss}(\bm{s}^{pred}_{seq}, \bm{u}^{opt}_{seq}) = &||\bm{z}^{pred}_{seq}-\bm{z}^{ref}_{seq}||_{2} + w_{1}||\bm{f}^{pred}_{3:T}-\bm{f}^{pred}_{2:T-1}||_{2}\nonumber\\
    &+ w_{2}||\bm{l}^{pred}_{3:T}-\bm{l}^{pred}_{2:T-1}||_{2} + w_{3}||\bm{u}^{opt}_{seq}||_{2} \label{eq:balance-loss}
  \end{align}
  where $w_{\{1, 2, 3\}}$ denotes the constant weight.
  (c) shows the results of five transitions of $z_x$ after an external force of 30 N is applied to the waist link for 0.2 s in the simulation.
  It can be seen that the convergence of $z_x$ to the external force becomes faster by the balance control.
  In addition, the average $E_z$ of the total error of $|z_x|$ for 6 seconds (for 30 steps) drops from 0.816 to 0.617 with the balance control.
  Note that the usual PD control is slow in convergence due to the delay caused by the flexibility of the body.
  The result was $E_z=0.791$ when setting the PD gains as (0.03, 0.1), and it was not much different from the result without any control, even after the parameters are manually tuned.
  (d) shows the results of five transitions of $z_x$ after applying 15 N force to the waist link and then releasing it in the actual robot.
  It can be seen that the convergence of $z_x$ to the external force becomes faster by the balance control.
  Also, $E_z$ dropped from 0.349 to 0.296 with the balance control.
}%
{%
  本研究はモデル化困難性として, 全身筋骨格ヒューマノイドMusashi \cite{kawaharazuka2019musashi}のバランスモデルを扱う.
  また, 逐次的モデル変化として, モデル内に含まれない靴の変化や上半身姿勢の変化等の身体状態変化を扱う.
  感覚と運動については, $\bm{s}=\{\bm{z}, \bm{f}, \bm{l}\}$, $\bm{u}=\bm{l}^{ref}$とした.
  なお, \figref{figure:balance-control}の(a)に示すように, $\bm{z}$はzero moment point (zmp) $\in\mathbb{R}^{2}$ ($\{x, y\}$方向を$\{z_x, z_y\}$と表す), $\{\bm{f}, \bm{l}, \bm{l}^{ref}\}$は両足首に関係する筋の\{筋張力, 筋長, 指令筋長\} $\in\mathbb{R}^{12}$である.
  本実験は, 柔軟身体においてその身体状態の変化に適応しながらヒューマノイドのバランス制御を行う点で難しい.
  シミュレーションと実機について実験を行う.
  シミュレーションについては身体状態変化として, 腰関節のピッチ軸角度$\theta_{s-p}$とキャリブレーションのズレを表現した足首ピッチ関節角度のオフセット$\theta^{offset}_{a-p}$を扱う.
  身体状態を$\theta_{s-p}=\{-5.0, 0.0, 5.0\}$ [deg], $\theta^{offset}_{a-p}=\{-5.0, 0.0, 5.0\}$ [deg]の組み合わせである9種類に変化させながらデータを取る.
  PBを2次元として, 約300ステップのデータを9回, 計約2700ステップのデータを用いて学習している.
  この際, $L_{1}=0.045$, $L_{2}=0.018$, $L_{3}=0.047$, $L_{4}=0.296$となり, $L_{thre}=0.2$と設定することで, $\bm{y}_{4}=\bm{u}$が出力から削除されSTMとなった.
  最終的な訓練時の推論誤差は0.051であった.
  また, 実機では身体状態変化として, \figref{figure:balance-control}の(d)うちのどの靴を使うかと, 上半身の姿勢変化を扱う.
  靴を\{Hard-Bare, Soft-Pink, Soft-Navy\}の3種類, 上半身姿勢を4種類の計12種類に変化させながら(Hard-Whiteは学習には用いない)データを取る.
  データを取る際は倒れない範囲でロボットを動作させる必要があるため, ランダムな制御入力の変位を徐々に増やし, かつそのランダム幅を周期的に変化させながらデータを収集することで, バランス制御に有用なデータを取得する.
  PBを2次元として, 約300ステップのデータを12回, 計約3600ステップのデータを用いて学習しており, 訓練時の推論誤差は0.074であった.


  \figref{figure:balance-control}の(b)にはシミュレーションにおいて学習されたPB $\bm{p}_{k}$の配置を示すが, $\theta_{s-p}$と$\theta^{offset}_{a-p}$の軸に沿って規則的にPBが自己組織化していることがわかる.
  これは実機においても同様で, \figref{figure:balance-control}の(e)には実機において学習されたPB $\bm{p}_{k}$の配置を示すが, 靴ごとに$\bm{p}_{k}$がグルーピングされ, PBの空間が自己組織化している.
  (b)と(e)の両者についてオンライン学習を実行すると, 現在の$\bm{p}$が``Target''となる現在の身体状態において学習された$\bm{p}_{k}$に徐々に近づいていくことがわかる.
  また, 学習時に用いなかったHard-WhiteのPBはHard-Bareの上部に配置されたが, これはHard-WhiteとHard-Bare, Soft-NavyとSoft-Pinkがそれぞれ似た靴底の固さを持つという性質を反映していると考えられる.
  この際の平均ステップ数は280であり, 約56秒で適応することがわかる.

  \figref{figure:balance-control}の(c)と(f)にはDPMPBを使ったバランス制御実験の結果を示す.
  PBをオンライン学習したうえで, \secref{subsec:control}における$h_{loss}$を以下のように設定することで, バランス制御を行うことができる.
  \begin{align}
    h_{loss}(\bm{s}^{pred}_{seq}, \bm{u}^{opt}_{seq}) = &||\bm{z}^{pred}_{seq}-\bm{z}^{ref}_{seq}||_{2} + w_{1}||\bm{f}^{pred}_{3:T}-\bm{f}^{pred}_{2:T-1}||_{2}\nonumber\\
    &+ w_{2}||\bm{l}^{pred}_{3:T}-\bm{l}^{pred}_{2:T-1}||_{2} + w_{3}||\bm{u}^{opt}_{seq}||_{2} \label{eq:balance-loss}
  \end{align}
  ここで, $w_{\{1, 2, 3\}}$は重みの係数を表す.
  筋張力と筋長の速度, 指令筋長を最小化しつつ, zmpを指令値へと近づける制御となる.
  (c)はシミュレーションにおいて, 腰リンクに対して30Nの外力を0.2秒間加えた後の$z_x$の遷移を5回検証した結果である.
  バランス制御により外力に対する$z_x$の収束がより速くなることがわかる.
  また, 6秒間(30ステップ分)の誤差$|z_x|$の合計値の平均$E_z$は, 制御の有無で0.816から0.617へと下がった.
  なお, 通常のPD制御は身体の柔軟性による遅れから収束を早くすることは難しく, 手動で調整したPゲインを0.03, Dゲインを0.1とした制御でも$E_z=0.791$と, 制御をしない場合と結果はあまり変わらなかった.
  (d)は同様に実機において, 腰リンクに対して15 Nの力をかけてからこれを離した後の$z_x$の遷移を5回検証した結果である.
  バランス制御により外力に対する$z_x$の収束がより速くなることがわかる.
  また, 制御の有無で$E_z$は0.349から0.296へと下がった.
}%

\begin{figure}[t]
  \centering
  \includegraphics[width=0.95\columnwidth]{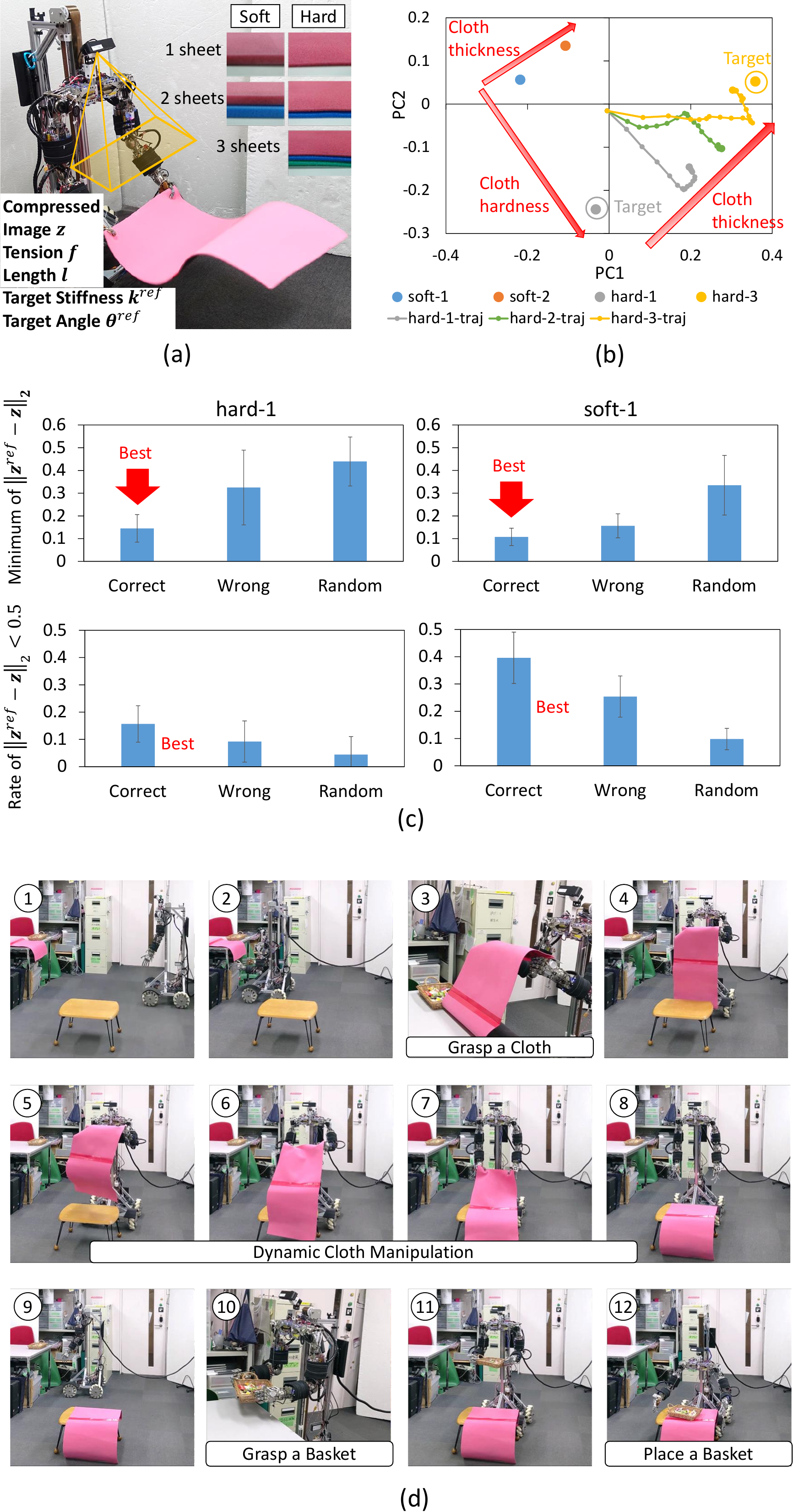}
  \vspace{-1.0ex}
  \caption{Experiment of cloth manipulation considering material change \cite{kawaharazuka2022cloth}. (a) shows the sensors and actuators of the musculoskeletal wheeled robot Musashi-W and the various cloths used in this experiment, (b) shows the trained parametric bias and its trajectory when conducting online update of PB, (c) shows the minimum value of $||\bm{z}^{ref}-\bm{z}||_{2}$ and the rate of $||\bm{z}^{ref}-\bm{z}||_{2}<0.5$ when conducting cloth manipulation with Correct / Wrong / Random settings, and (d) shows an integrated table setting experiment using the developed dynamic cloth manipulation.}
  \label{figure:cloth-manipulation}
  \vspace{-3.0ex}
\end{figure}

\subsection{Dynamic Cloth Manipulation Considering Material Change} \label{subsec:cloth-manipulation}
\switchlanguage%
{%
  In this section, we deal with a dynamic cloth manipulation model by a musculoskeletal wheeled robot Musashi-W as a modeling difficulty.
  As a temporal model change, we deal with the change of cloth material.
  For sensor state and control input, we set $\bm{s}=\{\bm{z}, \bm{f}, \bm{l}\}$ and $\bm{u}=\{\bm{\theta}^{ref}, \bm{k}^{ref}\}$.
  Note that, as shown in (a) of \figref{figure:cloth-manipulation}, $\bm{z}$ is the image compressed by AutoEncoder, $\{\bm{f}, \bm{l}\}$ is \{muscle tension, muscle length\} of the muscles related to both arms $\in\mathbb{R}^{20}$ (5 DOFs of the shoulder and elbow are used), $\bm{\theta}^{ref}$ is the target joint angle of the arm in the sagittal plane $\in\mathbb{R}^{2}$ (the pitch joints of the shoulder and elbow, the same for both arms), and $\bm{k}^{ref}$ is the target value for the stiffness of the arm $\in\mathbb{R}^{1}$.
  The cloth is composed of soft or hard foam sheets and is varied by stacking \{1, 2, 3\} sheets (denoted as \{soft-1, soft-2, hard-1, hard-2, hard-3\}).
  This experiment is difficult in that it is necessary for a robot with a flexible body to dynamically manipulate a flexible object, taking into account the change of its material.
  We collected data from random control inputs and human operations using GUI while changing the cloth material to \{soft-1, soft-2, hard-1, hard-3\}.
  DPMPB with 2-dimensional PB was trained using 16 datasets with about 400 steps each (about 6400 steps in total).
  Here, $L_{1}=0.122$, $L_{2}=0.193$, $L_{3}=0.037$, $L_{4}=0.233$, and $L_{5}=0.288$, and since $\bm{y}_{\{4, 5\}}=\bm{u}$ is removed from the output by setting $L_{thre}=0.2$, the network structure is STM.
  The final loss when training was 0.082.

  In (b) of \figref{figure:cloth-manipulation}, we show the arrangement of the trained PBs $\bm{p}_{k}$.
  We can see that the space of PB is neatly self-organized according to the hardness and thickness of the cloth.
  The trajectory of PB, represented as ``-traj'' in (b), shows that the current $\bm{p}$ approaches $\bm{p}_{k}$ trained in the current cloth when the online learning of PB is performed.
  For hard-2, which is not used for training, PB is at about the middle of hard-1 and hard-3.
  Here, the average number of data steps used for online update is 120, which indicates that the adaptation takes about 24 seconds.

  In (c) of \figref{figure:cloth-manipulation}, we show the results of dynamic cloth manipulation experiment using DPMPB.
  After updating PB online, we can transition the cloth to the desired state by setting $h_{loss}$ in \secref{subsec:control} as follows,
  \begin{align}
    h_{loss}(\bm{s}^{pred}_{seq}) = ||\bm{m}_{t}\otimes(\bm{z}^{ref}_{seq}-\bm{z}^{pred}_{seq})||_{2} + w_{1}||\bm{f}^{pred}_{seq}||_{2} \label{eq:loss-detailed}
  \end{align}
  where $\bm{z}^{ref}$ is the compressed latent value of the given target image and $w_{1}$ is the constant weight.
  This control realizes the target image while suppressing the muscle tension.
  Note that $\bm{m}_{t}$ ($\in \{0, 1\}^{N^{control}_{step}}$) is a vector in which 1 appears at every $N^{control}_{periodic}$ step and is otherwise 0 ($N^{control}_{periodic}$ is a constant value).
  At each step, the vector is shifted to the left and 0 or 1 is inserted from the right according to $N^{control}_{periodic}$.
  This makes it possible to make the cloth state $\bm{z}$ closer to the target value for each $N^{control}_{periodic}$ step, and enables the control to handle the dynamic state of the cloth that can be realized only for a moment.
  (c) shows the minimum value of $||\bm{z}^{ref}-\bm{z}||_{2}$ and the ratio of $||\bm{z}^{ref}-\bm{z}||_{2}<0.5$, which refers to the state where the current image is close to the target image.
  In addition, the following cases are compared: when PB is correctly set to the value of hard-1 for hard-1 and soft-1 for soft-1 (Correct), when PB is incorrectly set to the value of soft-1 for hard-1 and hard-1 for soft-1 (Wrong), and when the robot is moved randomly (Random).
  The minimum value is smaller in the order of Correct, Wrong, and Random, and the ratio of $||\bm{z}^{ref}-\bm{z}||_{2}<0.5$ is larger in the order of Correct, Wrong, and Random, indicating that the target state can be realized accurately by the control, and that the recognition of PB is important for this control.

  In (d) of \figref{figure:cloth-manipulation}, we show a series of table-setting experiments using this dynamic cloth manipulation model.
  Musashi-W picked up a cloth, laid it on the table as a tablecloth using this method, and placed a basket of sweets on the cloth.
}%
{%
  本研究はモデル化困難性として, 筋骨格ヒューマノイドMusashi-Wによる柔軟布操作モデルを扱う.
  また, 逐次的モデル変化として, 布の素材変化を扱う.
  感覚と運動については, $\bm{s}=\{\bm{z}, \bm{f}, \bm{l}\}$, $\bm{u}=\{\bm{\theta}^{ref}, \bm{k}^{ref}\}$とした.
  なお, \figref{figure:cloth-manipulation}の(a)に示すように, $\bm{z}$はAutoEncoderにより圧縮された画像, $\{\bm{f}, \bm{l}\}$は両腕(肩と肘の5自由度)に関わる筋の\{筋張力, 筋長\} $\in\mathbb{R}^{20}$, $\bm{\theta}^{ref}$は矢状面の腕の指令関節角度 $\in\mathbb{R}^{2}$ (肩と肘のピッチの2自由度, 両腕で同じ), $\bm{k}^{ref}$は腕の剛性を表す指令値 $\in\mathbb{R}^{1}$である.
  布はSoftまたはHardな発泡性のシート(実際の布はMusashi-Wの腕の速度では扱うことができなかった)を\{1, 2, 3\}枚に重ねて変化させる(\{soft-1, soft-2, hard-1, hard-2, hard-3\}と表す).
  本実験は, 柔軟身体を持つロボットが柔軟な物体を, その素材変化まで考慮しながら動的に操作する必要がある点で難しい.
  布の素材を\{soft-1, soft-2, hard-1, hard-3\}に変化させながらランダム動作とGUIによる人間の操作によりデータを収集する.
  PBを2次元として, 約400ステップのデータを16回, 計約6400ステップのデータを用いて学習している.
  この際, $L_{1}=0.122$, $L_{2}=0.193$, $L_{3}=0.037$, $L_{4}=0.233$, $L_{5}=0.288$となり, $L_{thre}=0.2$と設定することで, $\bm{y}_{\{4, 5\}}=\bm{u}$が出力から削除されSTMとなった.
  $L_{2}$もそれなりに大きいため$\bm{f}$の予測は難しいことが分かるが, 制御において筋張力最小化を行うため, 本研究ではこれを$\bm{s}$に含めた.
  なお, 最終的な訓練時の推論誤差は0.082であった.


  \figref{figure:cloth-manipulation}の(b)には学習されたPB $\bm{p}_{k}$の配置を示す.
  布の固さと枚数に応じて規則的にPBの空間が自己組織化していることがわかる.
  また, \figref{figure:cloth-manipulation}に``-traj''として表したPBの軌跡から, PBのオンライン学習を実行すると, 現在の$\bm{p}$が現在の布において学習された$\bm{p}_{k}$周辺に近づいていくことがわかる.
  学習に用いていないhard-2については, hard-1とhard-3の中間程度にPBが配置された.
  この際の平均ステップ数は120であり, 約24秒で適応することがわかる.

  \figref{figure:cloth-manipulation}の(c)にはDPMPBを使った動的布操作実験の結果を示す.
  PBをオンライン学習したうえで, \secref{subsec:control}における$h_{loss}$を以下のように設定することで, 布を所望の状態に遷移させることができる.
  \begin{align}
    h_{loss}(\bm{s}^{pred}_{seq}) = ||\bm{m}_{t}\otimes(\bm{z}^{ref}_{seq}-\bm{z}^{pred}_{seq})||_{2} + w_{1}||\bm{f}^{pred}_{seq}||_{2} \label{eq:loss-detailed}
  \end{align}
  ここで, $\bm{z}^{ref}$は与えた指令画像の圧縮表現, $w_{1}$は重みの係数を表す.
  筋張力を抑えつつ, 指令画像を実現する制御となる.
  なお, $\bm{m}_{t}$ ($\in \{0, 1\}^{N^{control}_{step}}$)は設定した$N^{control}_{periodic}$ステップごとに1が出現し, その他は0のベクトルである.
  一ステップごとに左へシフトし, 右からは$N^{control}_{periodic}$に応じて0または1が挿入される.
  これにより, $N^{control}_{periodic}$ごとに布の状態$\bm{z}$を指令値に近づけることができ, 周期的かつ, 一瞬しか実現できない布の動的な状態を扱う制御が可能となる.
  (c)はhard-1とsoft-1の2つの布素材について, 25秒間の実験を5回ずつ試行した際の, $||\bm{z}^{ref}-\bm{z}||_{2}$の最小値と$||\bm{z}^{ref}-\bm{z}||_{2}<0.5$ (現在画像が指令値に近い状態)となる割合を示している.
  また, 制御の際にPBがhard-1にはhard-1の, soft-1にはsoft-1の値に正しく設定した状態(Correct), PBがhard-1にはsoft-1の, soft-1にはhard-1の値に間違って設定した状態(Wrong), ランダムに動作した場合(None)を比較している.
  最小値はCorrect, Wrong, Randomの順で小さく, 割合はCorrect, Wrong, Randomの順で大きいことがわかり, 制御により指令状態が正確に実現できること, また, この制御はPBの認識が重要であることがわかった.

  \figref{figure:cloth-manipulation}の(d)には, この動的柔軟布操作モデルを使った一連のテーブルセッティング実験を示す.
  Musashi-Wが布を手に取り, 本手法によりテーブルの上にテーブルクロスとしてこれを敷き, その上にお菓子の入ったかごを載せるという一連の動作に成功した.
}%

\section{Discussion} \label{sec:discussion}
\switchlanguage%
{%
  We summarize the results obtained from our experiments.
  In this study, we found that DPMPB can learn a dynamics model between a flexible hand and objects, a visual feedback model of a low-rigidity body, a probabilistic relationship between a wheeled base and a floor, an imitation model of human motion, a balance model of a musculoskeletal humanoid, and a dynamic manipulation model of flexible cloth.
  DPMPB can learn not only the dynamics inside the robot body, but also the relationship it has with tools, objects, and environment.
  The parametric bias can embed differences in grasped objects, robot configurations, floor friction, human motion styles, shoes and cloth materials.
  DPMPB can recognize and adapt to the current state by updating only the parametric bias so that the network prediction matches the current state.
  It is also possible to recognize objects and states that are not used in training as intuitively correct dynamics in the space of PB.
  In the case of control using STM, the desired behavior can be achieved by constructing a loss function with a combination of minimization and maximization of a value, minimization of the error with a certain target value, and minimization of the change from the value of the previous time step.
  As an application of DPMPB, it was found that the stabilization control to minimize variance is possible by introducing the mean-variance representation to the network output, and that it is possible not only to adapt to the current state but also to actively change the motion style depending on the update rule of parametric bias in imitation learning.
  A series of behaviors using DPMPB is also possible, and DPMPB is expected to realize further behaviors to overcome modeling difficulties and temporal model changes.
}%
{%
  実験から得られた結果をまとめる.
  本研究では柔軟ハンドの動的モデルや低剛性身体の視覚モデル, 台車と床の確率的関係, 人間の動作模倣, 筋骨格型のバランスモデル, 柔軟布操作モデルをDPMPBにより学習可能であることがわかった.
  身体内部だけではなく, 身体から先の道具や対象物体との関係, 環境との関係までを学習することが可能である.
  Parametric Biasは把持物体の違いやロボット状態の違い, 床摩擦や人間の動作スタイル, 靴や布素材の違いを埋め込むことが可能であり, ネットワークの予測を現在状態に合致させるようにParametric Biasのみ更新することで, 現在状態を認識し適応することができる.
  また, 学習時に用いていない物体や状態についても, PBの空間で直感的に正しいダイナミクスとして認識することが可能である.
  STMによる制御の際は, 損失関数を状態値の最小化, 最大化, 速度最小化, 指令値との合致等の組み合わせにより構成することで, 所望の動作を実現することが可能であった.
  DPMPBの応用として, ネットワーク出力に平均分散表現を導入することで分散を最小化する安定化制御が可能であること, Parametric Biasの更新則次第で現在状態に適応するだけではなく能動的に動作スタイルを変更することも可能であることがわかった.
  このDPMPBで構成されたネットワークを利用した一連の動作も可能であり, ロボットのモデル化困難性と逐次的モデル変化を攻略したさらなる行動実現が期待される.
}%

\subsection{Limitations} \label{subsec:limitations}
\switchlanguage%
{%
  There are three main limitations in this study.
  First, the control period cannot be increased.
  The basic control period of the experiments on STM presented so far is 5 Hz, which is not fast.
  This is because the optimization process using iterative backpropagation and gradient descent methods in the control takes a long time.
  Note that the control period of CTM can be increased because the control input can be calculated from only the forward propagation.
  With the current network configuration, forward propagation takes about 10 msec and backward propagation takes about 40 msec.
  On the other hand, the optimization process can be accelerated by fixing the number of LSTM expansions in the control and learning the network as fully-connected layers \cite{kawaharazuka2019pedal}.
  We would like to continue development so that the robot can handle dynamic behaviors that require a control period of about 100 Hz.

  Second, in this study, tasks can only be controlled within the range that can be expressed by the loss function.
  Although the tasks handled in this study were successful, it is difficult to apply this method to tasks whose loss functions are more complex and difficult to be described by humans.
  It needs to be modified to be able to handle more abstract input/output variables.

  Third, we focus on the size of state/control space and data collection.
  In this study, we mainly trained the dynamics that is close to the robot body and easy to handle, but it is not suitable for motion planners with discrete states or task settings with large dimensions of control inputs and sensor states.
  It is also not suitable for tasks such as balance control or walking control where data collection itself is difficult.
  In order to handle discrete states and large state/control space in the real world, efficient data collection, definition of primitives, etc. would be needed.
  We also believe that it will be important to develop a method in which the robot itself determines whether the data collection is sufficient to perform the task and what the prediction error of the current network is.
  Although it should be theoretically possible to handle any kind of state transition, it is necessary to continue to verify the prediction accuracy and control performance when dealing with more sensors and nonlinearities.
}%
{%
  本研究の問題点は大きく分けて3つ存在する.
  まず, その制御周期を上げられない点が存在する.
  これまで示したSTMに関する実験の基本的な制御周期は5 Hzであり, とても速いとは言い難い.
  これは, 制御における誤差逆伝播と勾配降下法の繰り返しを利用した最適化計算に時間がかかるためである(CTMについては順伝播のみから計算可能であるため制御周期を上げることができる).
  現在のネットワーク構成では, 順伝播に約10 msec, 逆伝播に約40 msecの時間を有する.
  一方, 制御におけるLSTMの展開数を固定し, 全結合層として学習することで最適化操作を高速化可能である\cite{kawaharazuka2019pedal}.
  100 Hz程度の制御周期が必要なダイナミックな動作についても扱えるように今後の開発を続けたい.

  次に, 本研究は損失関数で表現できる範囲でしか制御を行うことができない点である.
  本研究で扱った実験には成功したものの, より損失関数が複雑で人間が記述することの難しいタスクへの適用は困難である.
  より抽象的な入出力変数を扱えるように修正する必要がある.

  最後に, 状態/制御空間の大きさとデータ収集が問題である.
  本研究はロボットの身体に近い扱いやすい部分について主に学習を行ったが, 離散的な状態を持つ動作計画器や, 制御入力や感覚の次元が大きなタスク設定には向かない.
  また, バランス制御や歩行制御のようなデータ収集自体が難しいタスクにも適していない.
  現実世界ではデータ収集の簡単さが重要になるが, 離散状態や大きな状態/制御空間等を扱うためには, より効率的なデータ収集とプリミティブ等の定義が必要になるケースが多くなる.
  また, そのデータ収集がタスクを行うのに十分であるか, ネットワークの予測誤差はどの程度かを, ロボット自身が判断しながらデータ収集する手法も今後重要であると考える.
  理論上どのような状態遷移であっても扱うことが可能であるはずだが, さらなるモーダルや非線形性を扱った際の予測精度・制御性能については今後も検証していく必要があると考える.
}%

\section{CONCLUSION} \label{sec:conclusion}
\switchlanguage%
{%
  In this study, we generalize a theory of deep predictive model learning with parametric bias, which can overcome modeling difficulties and temporal model changes in the relationship among the robot body, tools, target objects, and the environment.
  We constructed a predictive model using a neural network that overcomes modeling difficulties and an online update method of parametric bias that overcomes temporal model changes.
  In the network structure of the state transition model, the control input that makes the predicted state closer to the target state is calculated by repeating backpropagation and gradient descent methods for the network input, while in the network structure of the control transition model, the control input is calculated only from the forward propagation.
  The parametric bias, which can implicitly embed differences in dynamics into the network input, can be updated to make the predicted sensor state closer to the current sensor state in order to adapt to the current body and environment.
  Based on this predictive model learning, we have succeeded in the following tasks: grasping control and object recognition for a flexible hand, visual feedback for a low-rigidity robot, environmentally adaptive control with variance minimization for a wheeled robot, imitation learning for a musculoskeletal robot considering its motion style, balance control considering change in shoes for a full-body musculoskeletal humanoid, and dynamic cloth manipulation for a musculoskeletal wheeled robot considering cloth material change, and confirmed the effectiveness of DPMPB for various robot tasks.
  In the future, we would like to continue development so that the robot can autonomously collect data in the real world, acquire models of its body, tools, objects, and the environment, and perform various continuous tasks by coping with modeling difficulties and temporal model changes.
}%
{%
  本研究では, ロボットの身体-道具-対象物体-環境の間の相関関係におけるモデル化困難性と逐次的モデル変化を攻略可能な, Parametric Biasを含む深層予測モデル学習理論を一般化し定式化を行った.
  モデル化困難性を攻略するニューラルネットワークによる予測モデルの構築と, 逐次的モデル変化を攻略するParametric Biasのオンライン更新を行った.
  状態遷移モデル型の予測モデルではネットワーク入力に対する誤差逆伝播と勾配降下法を繰り返すことで予測状態を指令状態に近づける制御入力を計算し, 運動遷移モデル型の予測モデルでは順伝播のみから制御入力を計算する.
  暗黙的な状態変化をネットワーク入力に埋め込むことが可能なParametric Biasを, 予測状態と現在の感覚状態を近づけるように更新することで現在の身体や環境に適応することが可能になる.
  この予測モデル学習をもとに, 柔軟ハンドにおける把持制御と把持物体認識, 低剛性軸駆動型ロボットによる視覚フィードバック, 台車型ロボットの環境適応型分散最小化制御, 筋骨格型ロボットにおける動作スタイルを考慮した模倣学習, 筋骨格型台車ロボットの布素材変化を考慮した柔軟布操作, 全身筋骨格ヒューマノイドにおける靴の変化を考慮したバランス制御に成功し, 様々なロボット・タスクへの有効性を確認した.
  今後, 現実世界でロボットが自律的にデータを集め, 身体から道具,　対象物体, 環境までのモデルを獲得していき, モデル化困難性と逐次的モデル変化に対処して様々な連続タスクを行うことができるよう, 開発を継続していきたい.
}%

{
  \bibliographystyle{IEEEtran}
  \bibliography{main}
}

\end{document}